\RequirePackage{fix-cm}

\documentclass[smallextended]{svjour3}       

\smartqed  

\usepackage[nobottomtitles*]{titlesec}
\usepackage{graphicx}
\usepackage{todonotes}
\usepackage{multirow}
\usepackage{subcaption}
\usepackage{algorithmic}
\usepackage{graphics}
\usepackage{algorithm}
\usepackage{ stmaryrd }
\usepackage[toc,page]{appendix}
\usepackage{booktabs}
\usepackage{amsfonts}
\usepackage{xfrac}

\usepackage{natbib}
\setcitestyle{authoryear,open={(},close={)}}

\begin{document}

\title{Automated Algorithm Selection: from Feature-Based to Feature-Free Approaches }

\author{Mohamad Alissa \and
        Kevin Sim  \and
        Emma Hart
}

\institute{Mohamad Alissa \at
              Edinburgh Napier University, School of Computing, Edinburgh, UK \\
              \email{M.Alissa@napier.ac.uk} \\          
                ORCID: 0000-0002-9548-863X
           \and 
            Kevin Sim \at
              Edinburgh Napier University, School of Computing, Edinburgh, UK \\
              \email{K.Sim@napier.ac.uk}  
            \and
           Emma Hart \at
              Edinburgh Napier University, School of Computing, Edinburgh, UK \\
              \email{E.Hart@napier.ac.uk}  
}

\date{Received: date / Accepted: date}

\maketitle

\begin{abstract}
   We propose a novel technique for algorithm-selection, applicable to optimisation domains in which there is implicit sequential information encapsulated in the data, e.g., in online bin-packing. Specifically we train two types of recurrent neural networks to predict a packing heuristic in online bin-packing, selecting from four well-known heuristics. As input, the  RNN methods only use the sequence of item-sizes. This  contrasts to typical approaches to algorithm-selection which require a model to be trained using domain-specific instance features that need to be first derived from the input data. The RNN approaches are shown to be capable of achieving within 5\% of the oracle performance on between 80.88\% to 97.63\% of the instances, depending on the dataset. They are also shown to outperform classical machine learning models trained using derived features. {\color{black}Finally, we hypothesise that the proposed methods perform well when the instances exhibit some implicit structure that results in discriminatory performance with respect to a set of heuristics.} We test this hypothesis by generating fourteen new datasets with increasing levels of structure, and show that there is a critical threshold of structure required before algorithm-selection delivers benefit.

\end{abstract}

\keywords{Deep Learning, Machine Learning, Recurrent Neural Network, Algorithm Selection, Bin-Packing Problem, Feature-Free Approach.}

\maketitle

\section{Introduction}
\label{sec:intro}

\textit{Algorithm-selection} - the process of selecting the best algorithm to solve a given problem instance - is motivated by the potential to exploit the complementary performance of different algorithms on sets of diverse problem instances. 
However, determining the best-performing algorithm for an unseen instance has been shown to be a complex problem that has attracted much interest from researchers over the decades \citep{kotthoff2016algorithm,kerschke2018automated,Smith-Miles2009}. A common approach to tackling the Algorithm-Selection Problem (ASP) is to treat it as a classification problem where each instance is described in terms of a vector of hand-designed features, and an instance's class indicates the best performing algorithm. Although there have been a number of successful studies using this method, e.g. \citep{perez2004machine,kandanaarachchi2018normalization,collautti2013snnap,kerschke2019automated}, the task of identifying appropriate features that correlate to algorithm performance is far from trivial in many domains: in some domains, specifying features is not intuitive, and it can be difficult to create a sufficient number to train a model, while in others in which there are many features, it is necessary to invoke feature-selection methods in order to choose appropriate features \citep{kerschke2018leveraging,SMITHMILES201412} as the noisy and uninformative features prevent the selection techniques making intelligent decisions \citep{loreggia2016deep}.

Feature-design is even more complex in domains in which the data has sequential characteristics. For example, in online bin-packing \citep{lee1985simple,ramanan1989line} and online job-shop scheduling problems \citep{weckman2008neural,liu2009online}, items/tasks arrive in a stream (one at a time) and have to be packed/assigned to a container/machine  exactly in the sequence that they arrive. In such cases, it would be appropriate to derive features that capture the sequential information contained in the sequence in order to be informative, but deriving such features is even more challenging than in the cases mentioned above.

One solution to dealing with sequential data can be found in the field of deep learning, where the use of recurrent neural networks with Long-Short-Term Memory (LSTM) \citep{Hochreiter1997} or Gated Recurrent Units (GRU) \citep{cho2014learning} to  classify sequential data has become widespread in recent years; example applications include text  classification \citep{lee-dernoncourt-2016-sequential}, scene-labelling \citep{byeon2015scene} and time-series classification \citep{karim2017lstm}. Such networks directly use a sequence of data as input (e.g the size of the next item to be packed in bin-packing). In this sense they are `feature-free' in that it is not necessary to derive auxiliary features from the data to train the network. The addition of the LSTM/GRU to the network enables a model to learn the long-term context or  dependencies between symbols present in an input sequence, and also handles variable-length sequences of information.  Therefore, we propose that an RNN-LSTM or RNN-GRU could be used as a \textit{feature-free} classification technique to perform algorithm-selection in optimisation domains in which there is sequential data\footnote{in fact, it is possible to re-cast some optimisation problems that do not contain natural sequential information in sequence form; we return to this in the Conclusion.}. To be clear, in the context of this paper, the term \textit{feature-free} refers to the use of raw input data defining  a problem instance as input to a selector, where there is no pre-processing of data required or a need to define and derive features \emph{a-priori} from the problem data. This therefore addresses the associated issues outlined above. Also, we restrict the algorithm space to a set of deterministic heuristics, however, we use the phrase "algorithm-selection" as it is the familiar phrase of this problem in the literature.

In a recent conference paper \citep{alissa2019algorithm}, we described an initial implementation of an RNN-LSTM  to perform algorithm-selection in the field of 1D bin-packing, showing that it was able to outperform the Single-Best Solver (SBS) (i.e. the single heuristic that achieves the best performance over the instance set) on multiple datasets and achieving comparable performance to the Virtual-Best Solver (VBS), i.e. the oracle.  Our approach has subsequently been adopted by \citep{seiler2020deep} and modified to work in the TSP domain. Here we extend our previous work by proposing an additional neural architecture for prediction that uses gated-recurrent units. In addition, we compare the two RNN architectures to six different classical machine learning techniques that use derived features as input to provide better insight into the relative merits of recurrent vs classical networks. An extensive evaluation is conducted using five benchmark datasets.  In order to understand the conditions under which the proposed methods are likely to be useful, we then conduct a systematic investigation over fourteen newly created datasets that exhibit increasing levels of structure within the data, {\color{black}using a proxy for structure defined in terms of the performance difference between heuristics on the same instance.} This sheds new light on why some spaces are more likely to facilitate classification than others.

The contributions are as follows:
\begin{itemize}

  \item A novel feature-free algorithm-selection approach using a recurrent neural network with either long-short-term memory or gated recurrent units that avoids the need to identify features through training,  using only the sequential information defining a problem instance.
  \item An extensive comparison of the feature-free approaches to feature-based approaches, using a wide range of features as input to  six well-known classifiers, each tuned to ensure optimised performance.
  \item A systematic investigation of the relationship between classification performance and the level of structure in the dataset, using fourteen newly developed datasets that exhibit controllable levels of structure.

 \end{itemize}
 
Results show that on the five benchmark datasets, the feature-free approach significantly outperforms the best feature-based approaches, with classification accuracies across the five datasets  that are very close to the theoretical optimum. We further find that there is a threshold for `structure', below which algorithm-selection might be unnecessary. Given that many real-world problems are known to be structured \cite{Smith-Miles2009, hains2011revisiting} this underlines the need to develop better  algorithm-selection methods to be used in practice.

~\\

 The rest of the paper is organised as following, Section \ref{sec:relatedwork} explains the different approaches of dealing with ASP from the literature. A brief summary of Online 1D Bin-Packing Problem (1D-BPP) and the heuristics used in this study are presented in section \ref{sec:heuristics}. Section \ref{sec:Datasets} shows the problem instances we use in this research. We describe the Deep Learning and the Machine Learning models (including the features used) in section \ref{sec:dlMod}. This research methodology is explained in section \ref{sec:Methodology}. The results are presented in section \ref{sec:Results} followed by a systematic analysis across multiple datasets in section \ref{sec:ASP_Rand}.

\section{Related Work}
\label{sec:relatedwork}

\begin{figure}
\centering
\includegraphics[width=2.5in]{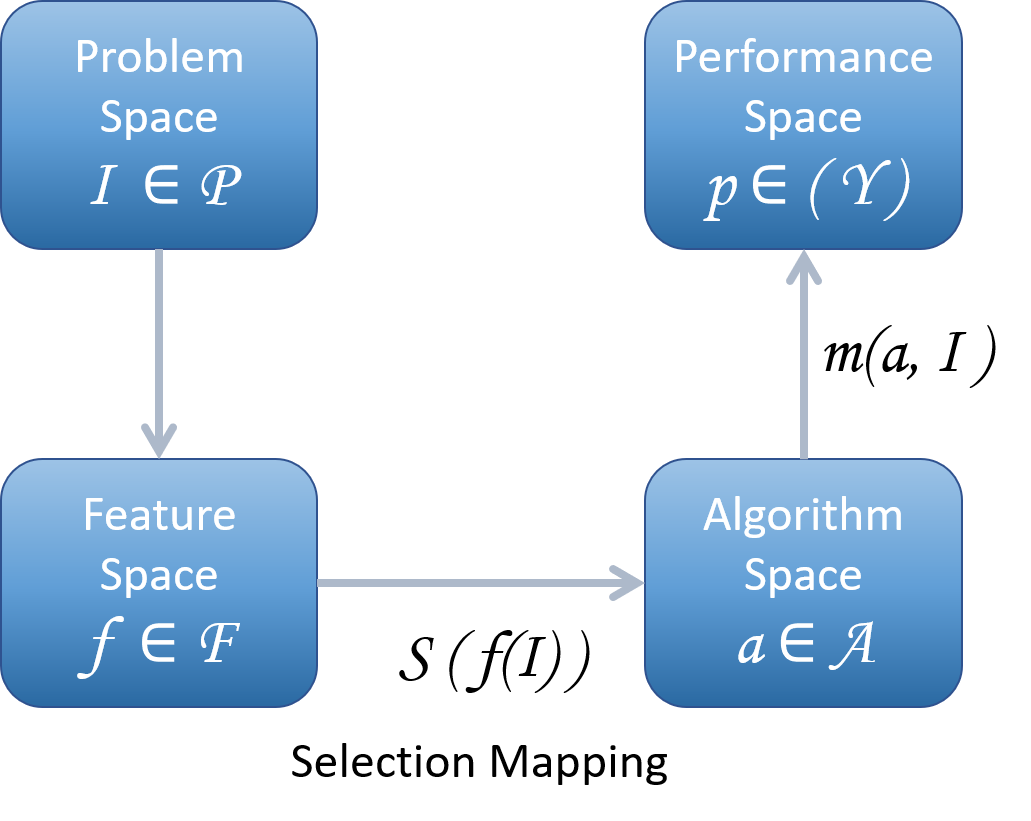}
\caption{Schematic of the Algorithm-Selection Problem \citep{Rice1976}}
\label{fig:riceASP}
\end{figure}

Originally formulated by \citet{Rice1976}, the per-instance Algorithm-Selection Problem (ASP) can be defined as:

\indent{\textit{"Given a set $I$ of instances of a problem $P$, a set $a = \{a_1, \ldots,a_n\}$ of algorithms for $P$ and a metric $m : a \times I \rightarrow$ $\mathbb{R}$  that measures the performance of any algorithm $a_j \in A$ on instance set $I$, construct a selector $S$ that maps any problem instance $i \in I$ to an algorithm $S(i) \in A$ such that the overall performance of $S$ on $I$ is optimal according to metric $m$."\citep{kerschke2018automated}}}

A schematic of the ASP is shown in Figure \ref{fig:riceASP} \citep{Rice1976}. In Rice's definition: 
\begin{itemize}
\item The problem space $\mathcal{P}$ represents a potentially infinite sized set of instances for the problem domain
\item The feature space $\mathcal{F}$ describes a set of characteristics derived using feature extraction from $\mathcal{P}$
\item The algorithm space $\mathcal{A}$ is the set of algorithms available for the problem domain
\item The performance space $\mathcal{Y}$ maps each algorithm to a set of performance metrics \citep{SMITHMILES201412}.
\end{itemize}

The objective is to identify a mapping between $\mathcal{P}$ and $\mathcal{A}$ that maximises $\mathcal{Y}$. In this paper, we restrict the algorithm space to a set of deterministic 1D-BPP heuristics.

For a finite set of problem instances $I$, a fixed set of heuristics $H$ and a single performance metric $m$, the Virtual Best Solver (VBS) is defined as a perfect mapping between $I$ and $H$. The Single Best Solver (SBS) is the heuristic $\in H$ that achieves the best performance over $I$. Although Rice's framework is a useful approach for describing ASP, it provides no advice about the mapping from problem space $\mathcal{P}$ to the feature space $\mathcal{F}$, and it clearly shows that the effectiveness of the algorithm-selection process for solving a particle problem domain relies on the quality of the problem's features \citep{Smith-Miles2011,smith2012measuring}. A comprehensive review of different approaches towards algorithm-selection can be found in a number of survey papers in this active area of research \citep{kerschke2018automated,kotthoff2016algorithm,Smith-Miles2009}. Some of the most relevant approaches are described here. 

\subsection{ASP with Feature-based Approaches}

One of the most common approaches to ASP is to identify features using expert knowledge,  and then train machine learning methods to predict the best performing algorithm(s) for an instance from its feature-profile \citep{kerschke2018automated}. However, identifying features that are significant in determining performance  is complex, usually requires hand-crafting \citep{Nudelman2004,Hutter2014, Smith-Miles2011,Pihera2014}, and often is not intuitive. Often the approach must also be combined with a \textit{feature-reduction} method to simplify learning, e.g Principal Component Analysis (PCA) \citep{SMITHMILES201412,lopez2013understanding}, and understand the correlations between features and algorithm performance.

\citet{cruz2012algorithm} have used meta-learning and hyper-heuristics to solve ASP in the domain of 1D-BPP. Their methodology relies on data collected from past experience to characterise algorithm performance and it is divided into three phases: initial training, prediction, and training with feedback. The output of the training phase is a trained model that relates the problem characteristics to the algorithms' performance --- this model is used to predict the best algorithm for a new given instance in the prediction phase. The new solved instances are then incorporated into the knowledge base to improve the selection quality. They used five deterministic heuristics and two non-deterministic algorithms with 1D-BPP. Three machine learning methods are compared --- Discriminant Analysis (DA) \citep{perez2004statistical}, a decision tree to build the selectors and  a Self-Organizing Map (SOM) \citep{haykin2009neural} to implement the selection system with feedback. Five features were used as input. Their method obtained 76\% accuracy with DA and 81\% accuracy with decision tree to select the best algorithm. Also, the accuracy increased from 78.8\% with initial-training up to 100\% when using SOM with feedback and the number of problem characteristics was the minimum.  

\citet{lopez2013understanding} also studied ASP in the packing domain using a wide range of 23 features and six heuristics within an evolutionary hyper-heuristic framework. They studied the correlation between the structure of 1D- and 2D-BPP instances and the performance of the solvers using PCA \citep{ringner2008principal} as a knowledge discovery method. Most of the used features are related to 2D-BPP and a subset of nine features, including means and standard deviation (std) of the item sizes, are considered that is strongly correlated with the heuristics performance after the feature-reduction. They analysed the distribution of feature values across the PCA map and their analysis suggested that there are indeed correlations between instance characteristics and heuristic performance. \citet{brownlee2018relating} have used ten BPP features that are related to the distribution of item sizes within each instance and performance features to analyse the relationship between the training data and automatic design of algorithms.  They investigated the distributions of values for features over the instances in benchmark sets, and how these distributions relate to the performance of algorithms built by automatic design of algorithms. They concluded that high variation in some of these features, including mean, standard deviation and maximum of item size, is a strong indicator for good fitting to the training instances and to achieve good performance for automatic design of algorithms.

A different type of approach was proposed by \citet{ross2002hyper} that could be used for ASP with constructive approaches to solution generation; rather than deriving features from the original description of an instance, a small number of features were derived from the current instance state each time a heuristic was applied. A learning classifier system XCS \citep{wilson1995classifier} was used to map a set of problem-states to specific heuristics. An approach that tries to avoid having to hand-craft good features was described in \citep{sim2012hyper} who evolve the\textit{ parameters} of a feature design method for 1D bin-packing problems to  that best improve the performance of k-Nearest Neighbours (KNN) classifier \citep{shalev2014understanding}.

Another ASP approach that does not explicitly  rely on feature identification and extraction was proposed by \citet{sim2015lifelong}. Here, a system continuously generates novel heuristics which are maintained in an ensemble, and samples multiple problem instances from the environment. Heuristics that "win" an instance (perform best) are maintained. This was shown to rapidly produce solutions and generalise over the problem space, but required a greedy method of actually selecting between generated heuristics and hence does not fit with the classical ASP definition.

\subsection{ASP with Streaming Problems}
Although feature-based approaches have been shown to work well in domains in which there is no sequential information associated with an instance description\footnote{although it could be argued that some sequential information is implicit in those approaches just mentioned that dynamically calculate problem state and use this to select heuristics \citep{sim2012hyper,ross2002hyper}.}, domains in which data arrives in a continual stream are more challenging. Statistical approaches to defining features for streaming data are complex,
and developing algorithm selectors to tackle streaming data poses considerable challenges due to potentially large streams, the fact that  the order of data points cannot be influenced and that the underlying distribution of the data points in the stream can change over time.  A recent survey article describing the state-of-the-art in algorithm-selection \citep{kerschke2019automated} highlighted a pressing need to develop automated algorithm-selection methods that are capable of learning in the context of streaming data. A supervised-learning approach was used by  \citet{van2018online, van2014algorithm} to predict which classifier performs best on a (sub)stream. Unsupervised learning approaches such as stream-clustering have been used to identify, track and update clusters over time \citep{carnein2017empirical,gong2017clustering}. 
However, due to the huge space of parameter and algorithm combinations, clear guidelines on how to set and adjust them over time are lacking \citep{mansalis2018evaluation, carnein2019optimizing,amini2014density}.

\subsection{ASP with Deep Learning Approaches}
Recently, deep learning algorithms have gained some traction in the ASP field  due their ability to learn from extremely large datasets in reasonable time. \citet{mao2017} proposed a heuristic performance predictor using a deep neural network trained on a large set of instances of variable sized 1D bin-packing problem using 16 features as input, grouped into item, box and cross features. Their prediction system has achieved up to 72\% validation accuracy to select the best performing heuristic that can generate a better quality bin-packing solution.  To eliminate the arduous task of manually designing features, \citet{loreggia2016deep} proposed a deep learning approach to automatically derive features in SAT and CSP domains assuming that any problem instance can be expressed as a text document. Unlike previous works e.g. \citep{SMITHMILES201412,lopez2013understanding} that derive features from features automatically using PCA, their approach automatically derives features  from a visual representation of the problem instances (i.e. converting the text files into grey-scale square images), which can be used to train a conventional neural network to predict the best solver for the instance. Although their approach obtained better results than the SBS, it was not able to outperform over the approaches that use regular manually crafted features.
Although concerned with learning an optimisation method rather than algorithm-selection, \citet{hu2017solving} used a deep reinforcement learning (a Pointer Network), with 3D-BPP to optimize the sequence of items to be packed into the bin by choosing the sequence, orientation and empty maximal space to pack cuboid shaped items. They claimed that their proposed method has obtained about 5\% improvement over a  well-designed heuristic.

As mentioned in the introduction, in a recent conference paper, \citet{seiler2020deep} adopted and adapted the LSTM approach  we proposed in \citep{alissa2019algorithm} to be applicable to the Euclidean TSP domain, also using an evolved (and balanced) dataset (1000 instances) with two TSP solvers. They compared a feature-based approach using four different classical ML classifiers to a feature-free approach using deep learning Convolutional Neural Networks (CNNs).
Due to the large TSP-related feature sets, they conducted a data analysis and automatic feature selection to choose adequate set of 15 most relevant features. Their results show that the feature-based approach improved over the SBS performance but still quite far away from the performance of the oracle-like VBS. The feature-free approach matches the performance of the quite complex classical ML approaches, despite being solely based on raw visual representation of the TSP instances. Although TSP is not an online or sequential problem, the work of \citet{seiler2020deep} borrows the key concept of our proposed method, i.e. that the raw data defining an instance can be used without modification as input to a selection algorithm.

The approach proposed in this paper differs substantially from the previous work just described in that it abandons the need to derive features from a dataset, circumventing the associated issues. In  contrast to some previous research which \textit{extends} Rice's diagram to encapsulate a broader agenda relating to the relative power of algorithms (e.g.  \citet{SMITHMILES201412}),  our proposed method shrinks Rice's diagram through bypassing the feature extraction block. Furthermore, as far as we are aware, it provides the first example of applying a recurrent-neural network as an algorithm-selector to data which has sequential characteristics. Although such networks have demonstrated ground-breaking performance on a variety of tasks that include image captioning, language translation and handwriting recognition \citep{lipton2015critical}, their applicability has not been exploited within the ASP domain.

\section{Online 1D-BPP: Definition and Heuristics}
\label{sec:heuristics}

The objective of the general 1D Bin-Packing Problem (1D-BPP) is to find a packing which minimises the number of containers, $b$, of fixed capacity $c$ required to accommodate a set of $n$ items with weights $\omega_j : j \in \{1, \ldots, n\}$ falling in the range $1 \leq \omega_j \leq c, \omega_j \in \mathbb{Z}^{+}$ 
whilst enforcing the constraint that the sum of weights in any bin does not exceed the bin capacity $c$.  The lower and upper bounds on $b$, $(b_l$ and $b_u)$ respectively, are given by Equation \ref{equation:BPP Bounds}. Any heuristic that does not return empty bins will produce, for a given problem instance, $p$, a solution using $b_p$ bins where $b_l \leq b_p \leq b_u$.

\begin{equation}
	\label{equation:BPP Bounds} 	
    b_{l} = \left \lceil \frac{1}{c} \sum \limits_{j = 1}^{n} \omega_j \right \rceil,~
	b_{u} = n	
\end{equation}

In \textit{online} bin-packing, items arrive in a stream, one at a time, and \textit{must} be packed in the order that they arrive.  In the specific version that we consider here, all items to be packed are known before packing starts (i.e. they constitute a fixed length batch) but the order that items in the batch are presented to the packing heuristics is fixed and cannot be changed. In contrast to other types of packing problem, the sequence cannot be re-ordered to find an ordering that provides an optimal packing with respect to a given heuristic. The function of the algorithm-selection method is therefore to select a heuristic to apply to pack the entire batch, considering the items in the fixed order given.

There have been numerous studies over the decades that have investigated the performance of simple approximation algorithms for the online variant of the BPP\citep{Johnson1974,DELORME20161}. We select 4 simple approximation algorithms from the literature specifically designed for this variation of the BPP  \citep{garey1981approximation} in order to evaluate the proposed algorithm-selection methods:

\begin{itemize}
 \let\labelitemi\labelitemii
  \item \textbf{First Fit (FF):} Places each item into the first feasible bin that will accommodate it.
  \item \textbf{Best Fit (BF):} Places each item into the feasible bin that minimises the residual space.
  \item \textbf{Worst Fit (WF):} Places each item into the feasible bin with the most available space.
  \item \textbf{Next Fit (NF):} Places each item into the current bin.  
\end{itemize}

For all the algorithms listed, if no feasible bin is available to accommodate the next item then it is placed into a newly opened bin. NF is different to the other 3 algorithms in that it only ever considers the most recently opened bin. If an item cannot fit in the current bin that bin is closed and removed from the problem. 

The performance of an algorithm $A$ on instance $\mathcal{I}$ is denoted by $A(\mathcal{I})$. $OPT(\mathcal{I})$ is the optimal solution for that instance. The worst-case performance ratio (WCPR) of $A$ is defined as the smallest real number $r(A) > 1$ such that $\frac{A(\mathcal{I})}{OPT (\mathcal{I})} \leq r(A)$ for all possible instances. The WCPR of NF is known to be 2 \citep{DELORME20161} and it was recently concluded after many theoretical studies that the WCPR of FF and BF is $\frac{17}{10}$\citep{Dosa2014}.

\section{Problem Instances}
\label{sec:Datasets}

We use a set of benchmark BPP instances that were first introduced in \citep{alissa2019algorithm}. These benchmarks consists of four balanced datasets: each dataset has 4000 instances, and contains exactly 1000 instances uniquely solved best by each of the four heuristics described in Section \ref{sec:heuristics}. The datasets were generated using an Evolutionary Algorithm which maximises the difference in a function $f$ between the target algorithm and the other algorithms used in the selection problem, where $f$ is Falkenauer's fitness function \citep{falkenauer1992genetic} given in equation \ref{eq:Falk} and are described in detail in \cite{alissa2019algorithm}.

\begin{equation}
	\label{eq:Falk} 	
    Fitness = \frac{1}{b} \sum_{i=1}^{b}({\frac{fill_i}{C}})^{k}
\end{equation}

Each instance in each dataset is labelled with the heuristic that provides the best result according to equation \ref{eq:Falk}. This metric is commonly used to gauge the quality of a solution produced by an bin-packing algorithm and returns a value between 0 and 1. In the original datasets described in \citep{alissa2019algorithm}, $k$ is fixed at 2. $C$ is the bin capacity which is fixed at 150, $fill_i$ is the sum of the item sizes in $bin_i$ and $b$ is the number of bins used.

A new dataset is created  by combining instances selected from all 4 datasets  just described (identified as DS5). This facilitates an investigation into whether the feature-based and feature-free models generalise across a mixed set of instances of different lengths with item weights drawn from different probability distributions and bounds. DS5 contains 4000 instances with 1000 instances selected from each dataset DS1-DS4. For each dataset, 250 instances were selected at random for each class (FF, BF, WF and NF), resulting in a balanced dataset containing equal numbers of instances from each class and each distribution.  Table \ref{tab:dsGen} provides a description of each dataset. These datasets are available for other researchers working in the field of ASP to compare approaches\footnote{https://github.com/Kevin-Sim/BPP}.

\begin{table}[htb]
  \centering
  \caption{Dataset Parameters. Bin Capacity is fixed at 150 \citep{alissa2019algorithm}}
    \begin{tabular}{cccc}
    \toprule
    Dataset & $n_{items}$ & Lower - Upper Bounds & Distribution $\mathcal{D}$\\
    \midrule
    DS1   & 120   & [40-60] & Gaussian \\
    DS2   & 120   & [20-100] & Uniform \\
    DS3   & 250   & [40-60] & Gaussian \\
    DS4   & 250   & [20-100] & Uniform \\
    DS5 &  (120,250) & [20-100, 40-60] & (Uniform, Gaussian) \\
    \bottomrule
    \end{tabular}%
  \label{tab:dsGen}%
\end{table}%

\section{Models: A Deep Learning Model and a set of Classical Machine Learning Models for Algorithm-Selection \label{sec:dlMod}} 

We have outlined above that conventional machine learning techniques used for the ASP tend to consider vectors of features as input to classical machine learning models. Candidate features typically describe spatial or statistical characteristics, with little consideration to ordering or sequential information describing an instance. On the one hand, this ignores potentially valuable sequential information that could improve algorithm-selection, while on the other,  limits applicability of standard approaches on streaming data where features need to be calculated dynamically.

We propose a model applicable to domains in which data has a fixed ordering that explicitly considers the ordering as input to an algorithm-selection technique, that uses a method borrowed from the deep learning literature.
Deep learning has achieved ground breaking results in applications where the input is formatted as time-series data or in domains where sequences have specific orderings but without any explicit notion of time \citep{lipton2015critical}.  Examples including video and image recognition, natural language processing, music generation and speech recognition \citep{graves2012supervised,Pouyanfar2018, skansi2018introduction}.

For comparison purposes, we compare results to conventional machine learning techniques that use feature-based input. Out of curiosity, we additionally evaluate the performance of two conventional machine learning models (Multi Layer Perceptron (MLP) and Random Forest (RF)) trained using the ordered list of item weights as input, i.e. without knowledge-driven feature extraction. The purpose of this is to investigate whether a conventional classifier can learn anything from feature-free input. Both models and their results are described in the end of section \ref{sec:Accuracy}.

\subsection{Deep Learning Feature-Free Model}
\begin{figure}[h]
\centering
\includegraphics[keepaspectratio,height=3in, width=3in]{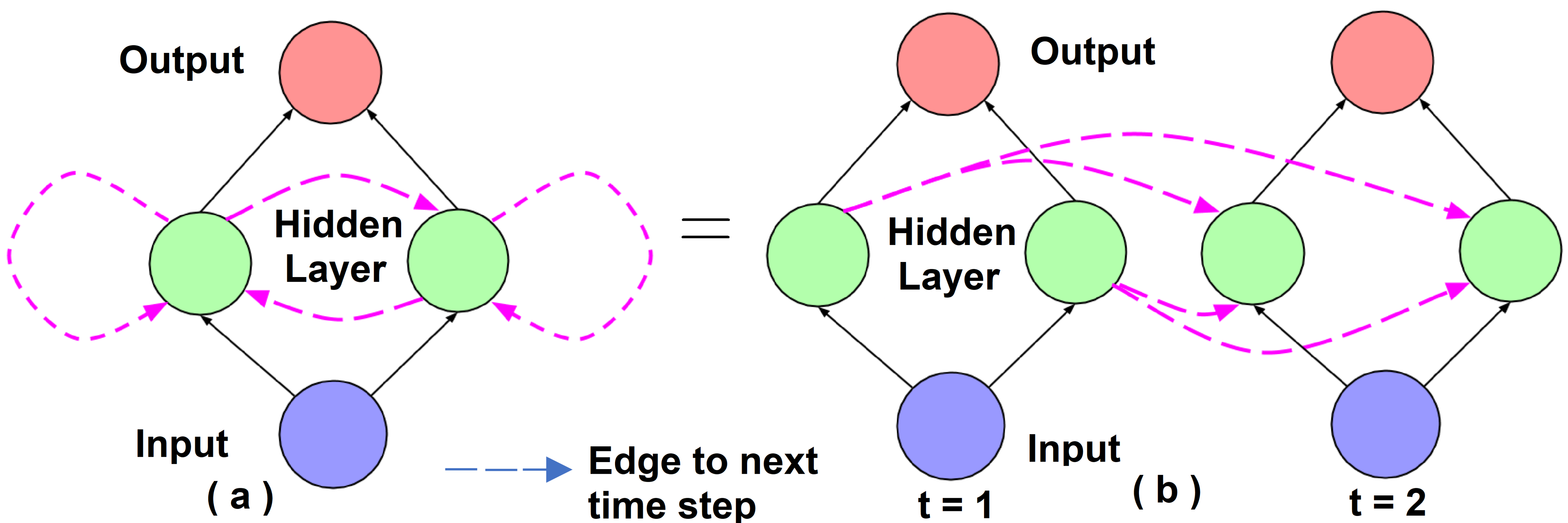}
\caption{(a) a Simple RNN and (b) an example of unfolded RNN with two time steps \citep{lipton2015critical}}
\label{fig:RNN}
\end{figure}

We select a Recurrent Neural-Network (RNN) as a deep learning method designed to learn from sequence data.
RNNs are one of the two most common architectures described under the umbrella term Deep Learning (DL). They differ from  Feedforward Neural Networks due to the presence of cyclic connections from each layers' output to the next layers input, with feedback loops returning to the previous layer (Fig \ref{fig:RNN}-a) \citep{wang2017origin,lipton2015critical}. This structure prevents traditional back-propagation being applied since there is not an end point where the back-propagation can stop. Instead, Back Propagation Through Time (BPTT) is applied: the RNN structure is unfolded to several neural networks with certain time steps and then the traditional back-propagation is applied to each one of them (Fig \ref{fig:RNN}-b) \citep{wang2017origin}. RNNs are specifically designed to learn from sequence data where sequential information explicit in the order of sequences is used to identify relationships between the data and the expected outputs from the network.

\begin{figure}[h]
\begin{minipage}[t]{0.49\linewidth}
\centering
\includegraphics[keepaspectratio,height=2.2in, width=2.2in]{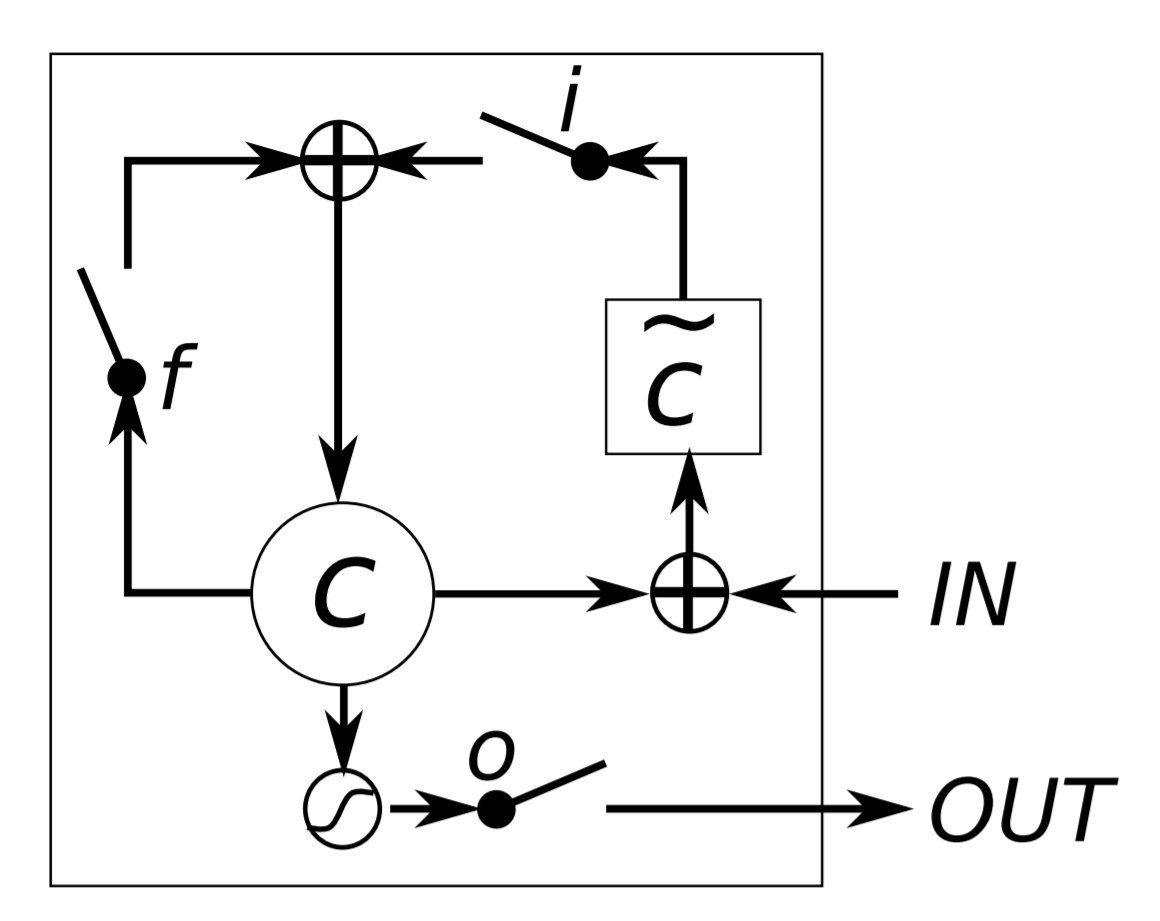}
\subcaption{ Long Short-Term Memory}
\label{fig:LSTM}
\end{minipage}
\begin{minipage}[t]{0.49\linewidth}
\centering
\includegraphics[keepaspectratio,height=2.2in, width=2.2in]{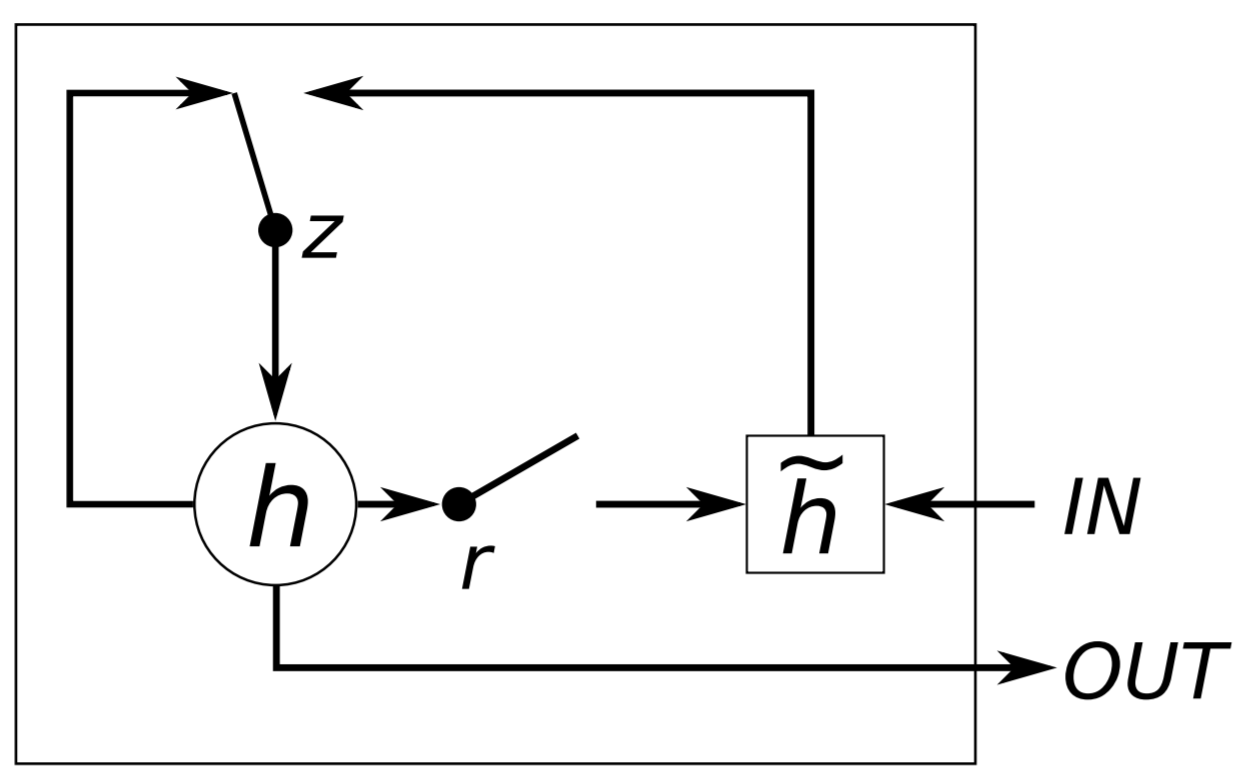}
\subcaption{Gated Recurrent Unit} \label{fig:GRU}
\end{minipage}

\caption{Graphical illustration of (a) LSTM, where \textit{i}, \textit{f} and \textit{o} are the input, forget
and output gates, respectively. \textit{c} and \textit{c˜} denote the memory cell and the new memory cell content. (b) Gated Recurrent Unit, where \textit{r} and \textit{z} are the reset and update gates, and \textit{h} and \textit{h˜} are the hidden stat and the new hidden stat, respectively \citep{chung2014empirical}.}
\end{figure}

In this study we use a specialised RNNs known as a Long Short-Term Memory (LSTM) \citep{Hochreiter1997} and Gated Recurrent Unit (GRU) \citep{cho2014learning} that have been shown to be efficient and effective in learning long-term dependencies from sequences of ordered data. LSTM neural networks incorporate additional gates that can retain, retrieve and forget information (i.e. the network states) over long periods of time \citep{graves2012supervised}. These gates are simply a combination of addition, multiplication and non-linear functions \citep{nielsen2015neural}. Three main gates are used in the LSTMs input, output and forget gates and three main states input, hidden and internal cell states. Basically, internal states are the ''memory'' of the LSTM block, hidden states represent values that come from the previous time step, and the input state is the result of the linear combination between the hidden state and the input of current time step. In the LSTM network, the classic neurons in the hidden layer are replaced by memory blocks. Fig \ref{fig:LSTM} shows that the LSTM's input comes from the network through the input gate and the only outputs from the LSTM to the rest of the network emanate from the output gate multiplication. The input gate determines how much of the new memory content is added to the memory cell, the output gate modulates how much of the internal state would be exposed to the external network (higher layers and the next time step), while the forget gate defines how much of the existing memory is forgotten. GRU is a recent variation on LSTM with only two gates, update and reset gates which decide what information should be passed to the output. The update gate decides how much of the past information would be passed  to the future while the reset gate determines how much to be forgotten. Also, GRU does not use the internal state and instead uses the hidden state to transfer information through the time steps \citep{chung2014empirical}.
A more comprehensive description of the rapidly expanding field of DL, which has many competing, but no prevalent architectures, is outwith  the scope of this study.

\subsection{Feature-based machine learning models}
We evaluate six well-known classical machine learning techniques \citep{shalev2014understanding} from the literature: Neural Network (NN); Decision Trees (DT); Random Forest (RF); Naive Bayes (NB); k-Nearest Neighbours (KNN) and Support Vector Machine (SVM). Each is used in its standard form, parameterised using grid-search and bayesian optimisation as described in section \ref{sec:Methodology}. The success of ASP methods that rely on feature extraction depend critically upon the chosen features \citep{smith2012measuring} and on the classification method used to learn the correlation between the features and best-performing heuristic.

\subsection{Definition of features}
\label{sec:features}

The traditional method of dealing with ASP is to hand-design features and then extract them from the instance data. Here, we use a set of features collated from multiple papers in the literature where algorithm-selection methods have been applied to bin-packing \citep{cruz2012algorithm,lopez2013understanding,brownlee2018relating}, discussed in section \ref{sec:relatedwork}. The features are given below:

\begin{enumerate}
    \item Mean item size divided by bin capacity $C$ \citep{brownlee2018relating,lopez2013understanding,cruz2012algorithm,mao2017};
    \item Standard Deviation (std) in the item sizes divided by bin capacity $C$ \citep{brownlee2018relating,lopez2013understanding,cruz2012algorithm,mao2017};
    \item Maximum item size divided by bin capacity $C$ \citep{brownlee2018relating,mao2017};
    \item Minimum item size divided by bin capacity $C$ \citep{brownlee2018relating,mao2017}; 
    \item Median item size divided by bin capacity $C$ \citep{brownlee2018relating};
    \item Maximum item size divided by minimum item size \citep{brownlee2018relating}; 
    \item The ratio of small items of size $\omega$ $\leq$ $C/4$ \citep{ross2002hyper};
    \item The ratio of medium items of size $C/4$ $\textless$ $\omega$ $\leq$ $C/3$ \citep{ross2002hyper};
    \item The ratio of large items of size $C/3$ $\textless$ $\omega$ $\leq$ $C/2$ \citep{ross2002hyper};
    \item The ratio of huge items of size $\omega$ $\textgreater$ $C/2$ \citep{ross2002hyper}.
\end{enumerate}

It should be clear that these features do not provide any information about the sequential characteristics implicit in the data-stream that describes each instance (i.e. the order of items to be packed). Rather, they describe statistical properties relating to the distribution of the item sizes.  The 10 features were extracted for each of 16,000 instances contained in DS(1-5) and used as input to the classical machine learning models.

\section{Methodology}
\label{sec:Methodology}
We use both the Keras\footnote{https://github.com/fchollet/keras.} and the Sklearn \footnote{https://github.com/scikit-learn/scikit-learn.} libraries to implement the models used in the DL and classical ML experiments. We use a Keras implementation of LSTM and GRU in ``sequence-to-one mode'' where input is an ordered list of item weights and output is a ``one-hot''  encoding using 4 bits to identify the best heuristic (1000 = BF, 0100 = FF, 0010 = NF, 0001 = WF).  The Sklearn library is used to implement the classical ML models, with the exception of the NN model which is implemented using Keras using a one-hot encoding output.  All the classical models take 10 features as input as defined in section \ref{sec:features}. Experiments are conducted on Google Colab\footnote{https://colab.research.google.com/notebooks/welcome.ipynb.} with Tensor Processing Unit (TPU) run-time used to execute the experiments. A preliminary empirical investigation was conducted to tune both the LSTM and NN architectures and hyper-parameters. The "Adam" optimiser \citep{kingma2014adam} was selected due to its reported accuracy, speed and low memory requirements.

Due to the time limitation and the large number of hyper-parameters for LSTM/GRU and NN, we undertook preliminary investigations to optimise the LSTM\footnote{GRU has not been tuned: the best LSTM hyper-parameters are used with GRU since LSTM and GRU are very similar.} and NN approaches. We used 300 learning iterations for LSTM and GRU experiments on DS1 and DS2 and 700 learning iterations for DS3, DS4 and DS5 since the longer instances were found to require more learning iterations. On the other hand, both grid search and bayesian optimisation are used to optimise the remaining traditional machine learning techniques to choose the best set of hyper-parameters for each ML model. Tables \ref{table:LSTM-NN-Param} and \ref{tab:MLParam} in the appendix show the range of hyper-parameters evaluated and the final selected parameters for each of the LSTM, GRU, NN and ML models.

We conduct independent experiments using each of the five datasets to train and test the LSTM, GRU and classical ML models to predict the best heuristic for each instance. Each dataset was split into training (80\%) and test (20\%) sets while maintaining the balance in size of each target class (each test set has 800 instances where 200 are solved best by each heuristic). Each model was trained using 10-fold cross validation using 10\% of the training instances to verify each fold and to evaluate the generalisation error. Then the models are retrained on the whole corresponding training set and tested on the test set comprising 800 unseen instances. Each trained model is evaluated on an unseen test set in terms of the four following aspects ( these aspects also described in details in the next section \ref{sec:Results}),
\begin{enumerate}
    \item Comparison of algorithm classification accuracy, i.e. the percentage of instances classified correctly.
    \item  Comparison of selectors' performance to SBS and VBS using Falkenauer's performance metric (Equation \ref{eq:Falk}).
    \item  Comparison of selectors' solution quality using the number of used bins $b'$ by the predicted heuristic.
    \item Evaluate the ability to generalise based on the above criteria with a wide range of different randomly generated datasets.
\end{enumerate}

In addition to evaluating accuracy, we also report the total number of bins in a solution, summed over all instances in a dataset, when the heuristic returned by the model for each instance is used to create the solution. As the number of bins per instance in a given dataset can vary widely, to avoid issues that occur when summing data that has different scales, we normalise the total number of bins according to equation \ref{eq:binPerc} and sum the normalised values.

\begin{equation}
	\label{eq:binPerc} 	
	Percentage Of Bins =\frac{b' - b}{b},  \text{where} \hspace{4pt}  b = \left \lceil \sum_{j=1}^{n}({\frac{\omega_j}{C}}) \right \rceil, \hspace{4pt} \text{and}  \hspace{4pt} b \leq b' \leq 2b
\end{equation}

As mentioned in section \ref{sec:heuristics}, as the worst-case performance ratio for the heuristics considered is equal to the double the lower bound, then  Equation ~\ref{eq:binPerc}  returns a value between 0 and 1. A Wilcoxon signed-rank test is used to evaluate significance in a pairwise fashion for all comparisons of Falkenauer's performance and bins.{\color{black} This statistical test is corrected for multiple comparisons with the Bonferroni method \citep{weisstein2004bonferroni}, i.e. the p-values have been multiplied by the number of comparisons and then compared against the confidence level 5\%}. A statistical testing is conducted to evaluate three hypotheses:
\begin{itemize}
        \item $\mathcal{H}_0(1)$: the LSTM/GRU and best ML method produce equal results with respect to a) Falkenauer's fitness metric b) total bins utilised.
    \item $\mathcal{H}_0(2)$: the LSTM/GRU and SBS produce equal results with respect to a) Falkenauer's fitness metric b) total bins utilised.
    \item $\mathcal{H}_0(3)$: the best ML method and SBS produce equal results with respect to a) Falkenauer's fitness metric b) total bins utilised.
\end{itemize}

\section{Results}
\label{sec:Results}
This section discusses in details the four aspects we use to evaluate each trained model on an unseen test set. 

\subsection{Accuracy of Algorithm-Selection}
\label{sec:Accuracy}

As mentioned in section \ref{sec:Methodology}, 10-fold cross validation is used to train the models taking 10\% of the training instances to verify each fold and to evaluate the generalisation error. Table \ref{tab:MLValtset} in the appendix shows results achieved on the \textit{validation sets} used during training for the DL models (LSTM and GRU) and all 6 ML models. Table \ref{tab:MLtestset} shows the results achieved on each test set with the model obtained from training on the full training set. We report the classification accuracy as an indicator of the LSTM's, GRU's and ML's predictive abilities. Using Wilcoxon signed-rank test, significance is calculated in a pairwise fashion and {\color{black} corrected for multiple comparisons with the Bonferroni method} between the DL techniques and VBS, SBS and best of the ML techniques for each dataset, and between the best ML technique and the SBS. p-values are shown in table \ref{tab:Pval-DS[1-5]} in the appendix. Table \ref{tab:ranking_perform} additionally shows the comparison between the LSTM, GRU, the best ML technique and the SBS over the different test sets.

\begin{table}[h]
\caption{Classification accuracy of the trained model in each experiment from the test set, values in \textit{italic} indicate the best ML results and values in \textbf{bold}  the  best overall result per dataset. Significance at the 5\% confidence level (using paired Wilcoxon signed-rank test and {\color{black} corrected for multiple comparisons with the Bonferroni method}) is indicated as follows: $\ast$ indicates that the best DL model performed significantly worse
than the VBS; $\diamond$ indicates that the DL model performed significantly better
than the SBS; $\mathsection$ indicates that the DL model performed significantly better
than the best ML model; $\bullet$ indicates that the best ML model performed significantly better than the SBS in terms of Falkenauer's Performance.}

\centering
\resizebox{\columnwidth}{!}{%
\begin{tabular}{ccc|cccccc}
\hline 
 &  \multicolumn{2}{c}{DL} & \multicolumn{6}{c}{ML}  \\
\hline 
DataSet & LSTM & GRU & NN & DT & RF & SVM & NB & KNN \\
\hline 
DS1 & $80.88\%^{\ast\diamond\mathsection}$ & $\textbf{83.00\%}^{\ast\diamond\mathsection}$ & $\textit{66.88\%}^{\bullet}$ & 64.38\% & 66.50\% & 65.75\% & 61.75\% & 64.38\% \\
\hline 
DS2 & $90.63\%^{\ast\diamond\mathsection}$ & $\textbf{91.88\%}^{\ast\diamond\mathsection}$ & 58\% & 55.50\% & $\textit{58.62\%}^{\bullet}$ & 58.50\% & 55.88\% & 54.37\% \\
\hline 
DS3 & $80.88\%^{\ast\diamond\mathsection}$ & $\textbf{82.88\%}^{\ast\diamond\mathsection}$ & 66.63\% & 63.24\% & 65.63\% & $\textit{66.88\%}^{\bullet}$ & 51.37\% & 64.50\% \\
\hline 
DS4 & $95.50\%^{\ast\diamond\mathsection}$ & $\textbf{96.38\%}^{\ast\diamond\mathsection}$ & $\textit{72.88\%}^{\bullet}$ & 66.88\% & 72.88\% & 72.50\% & 64.75\% & 66.00\% \\
\hline 
DS5 & $85.00\%^{\ast\diamond\mathsection}$ & $\textbf{86.13\%}^{\ast\diamond\mathsection}$ & 59.50\% & 60.25\% & $\textit{63.74\%}^{\bullet}$ & 55\% & 41.13\% & 56.75\% \\
\hline 
\end{tabular} }
\label{tab:MLtestset}
\end{table}

\begin{table}[]
\caption{The comparison between the LSTM, the GRU, the best ML technique and the SBS over the different test sets in terms of Falkenauer's Performance using the paired Wilcoxon Signed-Rank Test with 5\% confidence level {\color{black} corrected for multiple comparisons with the Bonferroni method}. The $\shortuparrow$ means the left-hand approach of the given pair has a better median; the $\shortdownarrow$ means the first approach's median is worst; $+$ indicates significance ($\ll$ 5\%) and $-$ indicates no significance ($\gg$ 5\%).
}
\centering
\label{tab:ranking_perform}
\begin{tabular}{ccccccc}
\hline
 & \multicolumn{6}{c}{Falkenauer's Performance}  \\
 \hline
 & LSTM-SBS & LSTM-ML & GRU-SBS & GRU-ML &  ML-SBS & LSTM-GRU  \\
 \hline
DS1 & $\shortuparrow +$ & $\shortuparrow +$ & $\shortuparrow +$ & $\shortuparrow +$ & $\shortuparrow+$ &  $\shortdownarrow-$ \\
\hline
DS2 & $\shortuparrow +$ & $\shortuparrow +$ & $\shortuparrow +$ & $\shortuparrow +$ & $\shortuparrow +$ & $\shortdownarrow-$ \\
\hline
DS3 & $\shortuparrow +$ & $\shortuparrow +$ & $\shortuparrow +$ & $\shortuparrow +$ & $\shortuparrow+$ & $\shortdownarrow-$ \\
\hline
DS4 & $\shortuparrow +$ & $\shortuparrow +$ & $\shortuparrow +$ & $\shortuparrow +$ & $\shortuparrow +$ & $\shortdownarrow-$ \\
\hline
DS5 & $\shortuparrow +$ & $\shortuparrow +$ & $\shortuparrow +$ & $\shortuparrow +$ &  $\shortuparrow +$ & $\shortdownarrow-$ \\
\hline
\end{tabular} 
\end{table}

Results obtained from deep learning LSTM and GRU approaches are significantly better than the results using classical ML techniques (i.e. $\mathcal{H}_0(1)$ is rejected).
We observe that the DL results from the validation and test sets on DS2 and DS4 are better than on the other datasets suggesting that \textit{the sequential} correlations are easier to find in the datasets comprised of item weights generated from a wider range of values following a uniform distribution. It might be that the instances that are evolved with items in the range [40-60] may have fewer distinct patterns (i.e. sequential information) than instances generated with items in the range [20-100], regardless of the length of the instances.
It is well known that problems with an average weight of $\frac{C}{3}$ are more difficult to solve \citep{falkenauer1992genetic} and it is interesting that problems with those characteristics are more difficult to classify using LSTM or GRU, i.e. instances from DS(1,3) with item weights generated from a narrow range of values [40,60] which is an average of one third of bin capacity 150. Although all the ML results show relatively poor performance compared to the DL methods, it does comparatively better, in the most cases, on the two longer datasets DS3 and DS4 than the shorter ones DS1, DS2 on the validation set, while this is true only for the comparison between DS2 and DS4 on the test set. It might be that the longer instances result in more distinct values for the extracted features,  hence increasing the ability of the ML techniques to classify correctly.

Although training the LSTM and GRU on the longer instances requires significantly more learning iterations before the models converge, it is interesting to note that for both distributions the results obtained by the LSTM and GRU on the longer instances (e.g. DS4) exceed those reported on the datasets with smaller numbers of items (e.g. DS2) on the validation set. We conjecture that the longer instances provide the DL models with more sequential information, hence increasing the ability to determine patterns in the item sequences. The ML results partially concur with the LSTM and GRU results in this respect, i.e. results on DS3 are more accurate than DS1 (apart from NB) and those for DS4 are more accurate than for DS2 on the validation set. For both DL and ML models, this is true only for the comparison between DS2 and DS4 on the test set. The results of the LSTM and GRU experiments conducted on the combined DS5 set show intermediate results with accuracy between that achieved on the experiments on instances with uniform distribution and those with Gaussian distribution. These models are able to generalise over instances sampled from all of the problem lengths and the different weight distributions investigated without any apparent loss of precision. In contrast, the ML experiments conducted on the combined DS5 sets show worse results than most of the other experiments. This demonstrates that the feature-based approach is weak in its ability to  generalise over instances with different characteristics and highlights its reliance  on the quality of the designed features, in contrast to the LSTM and GRU approaches.

Table \ref{tab:DL_ML_CM_DS5} presents confusion matrices of the LSTM, GRU and best ML techniques extracted from the experiments conducted on the test set DS5, the rest of the confusion matrices are shown in table \ref{tab:DL_ML_CM} in the appendix. In most cases, the DL models were most frequently confused when attempting to classify the sequences identified as being solved best by FF and BF. It is interesting to note that \citep{alissa2019algorithm} previously showed that these two algorithms are extremely close in terms of their two largest principal components in a space defined by a 4-d vector containing the performance metric of each of the 4 heuristics for each instance. Similarly, instances labelled as NF appear to be the easiest to identify and correspondingly are the most isolated in the performance-space. As noted previously in \citep{alissa2019algorithm}, it appears that the patterns shown by conducting a PCA of the performance space are correlated with the ability of LSTM to identify the best performing algorithm from the raw instance sequences. On the other hand, the ML models are frequently confused between FF and BF (similarly to LSTM and GRU), but the ML models are also confused between FF and WF. In terms of the accuracy to classify NF, ML results partly concur with the DL results, in that NF instances are easier to classify only in DS1 and 3.

\begin{table}
\caption{The Confusion Matrix of the LSTM, GRU and Best ML models in experiments on DS5 from the test set}
\label{tab:DL_ML_CM_DS5}
   \centering
    \begin{tabular}{c||cccc|cccc|cccc}
    \hline 
    \multicolumn{1}{c}{ } & \multicolumn{4}{c}{LSTM} & \multicolumn{4}{c}{GRU}  &  \multicolumn{4}{c}{RF} \\
    \hline 
    Heuristic & BF & FF & NF & \multicolumn{1}{c|}{WF} & BF & FF & NF & \multicolumn{1}{c|}{WF} & BF & FF & NF & WF \\
    \hline 
    BF & 164 & 32 & 0 &  4 & 163 & 36 & 0 & 1 & 137 & 47 & 0 & 16 \\
    \hline 
    FF & 45 & 137 & 1 &  17 & 38 & 147 & 1 & 14 & 43 & 100 & 11 & 46 \\
    \hline 
    NF & 0 & 0 & 195 &  5 & 0 & 0 & 195 & 5 & 9 & 26 & 144 & 21 \\
    \hline 
    WF & 4 & 6 & 6 &  184 & 2 & 10 & 4 & 184 & 11 & 40 & 20 & 129 \\
    \hline 
\end{tabular}
\end{table}

Purely out of our interest, we conducted a 10-fold cross validation on DS4 using only the item sizes information defining the original instances (i.e. without any features) directly supplying all item sizes at once to two of the classical ML techniques: the Multi Layer Perceptron (MLP) and the Random Forest (RF) classifiers.  The ML techniques  were used directly from Weka \citep{Frank2016} without altering any of the default parameters. The RF achieved 67.55\% accuracy. MLP was equally successful, achieving 67.08\% accuracy. Although better than expected, we found that a DL approach that has been designed to work with sequential data provides more informative results.

\subsection{Comparison to SBS and VBS}

In terms of Falkenauer’s fitness, for all experiments the DL selectors significantly improve on both the SBS (BF for all datasets) and the classical ML techniques (i.e. $\mathcal{H}_0(1,2)$ are rejected), and are very close to the VBS. Also, the best ML techniques improve over the SBS significantly (i.e. $\mathcal{H}_0(3)$ is rejected), P-values are shown in table \ref{tab:Pval-DS[1-5]} in the appendix.

Figure \ref{fig:DS_Histo_Falk} shows  cumulative distribution plots over the test sets of 800 instances from each DS(1-5) in terms of Falkenauer’s fitness: the plots show the percentage of instances that are solved with an distance \textit{d$_{p}$} of the oracle-like VBS (the perfect mapping) given the solver predicted by a model $m$. The distance $d$ is calculated as  \textit{d$_{p}$}= (VBS$_{Falkenauer’s fitness}$ $-$ Selector$_{Falkenauer’s fitness}$). Hence, \textit{d$_{p}$} $\geq$ 0 and \textit{d$_{p}$ = 0} represents the optimal.

 Results show that the RNN-LSTM/GRU solves between 80.88\%-97.63\% of the instances within 5\% of the VBS performance, compared to 50\%-62.2\% using the SBS. This represents a 30.88\% to 35.43\% improvement on DS(1-5). The best ML technique solves 67.13\%-78.63\% instances with 5\% performance difference in all the datasets which is 16.43\%-17.13\% improvement over the SBS. Using either LSTM or GRU, between 99.63\% and 100\% of the instances can be solved within 30\% of VBS for DS(1-5): in contrast, using the best ML technique,  this is only true on DS(1,3). From the figure, it is clear that the best ML technique can only solve 100\% of instances if we consider a performance difference of 40\%  in DS(2,4,5). It is interesting to note that although the SBS solves a lower percentage of instances than both the DL and the best ML technique if we consider a difference of  5\% of the VBS performance, it manages to solve all the instances at 20\% of the VBS in all the datasets. Although the ML techniques improve over the SBS in the most cases, they are not as good as the LSTM/GRU and are not as close to the VBS as the DL methods.

It is noticeable from the evolved instances in \citep{alissa2019algorithm} that the performance of the heuristics is skewed towards FF and BF, i.e. these heuristics are either the best or second best choice for all instances. This means that even if the selector mis-classified most of the NF and WF instances by choosing FF or BF, it will still achieve high performance.
 Therefore,  as well as comparing the two approaches based on the performance metric and  classification accuracy,  we additionally compare them in terms of the number of bins used to pack a set of items to get a clearer understanding of their relative performance.

\begin{figure}[]
\begin{minipage}[t]{0.45\linewidth}
\centering
\includegraphics[angle=0,scale=0.30]{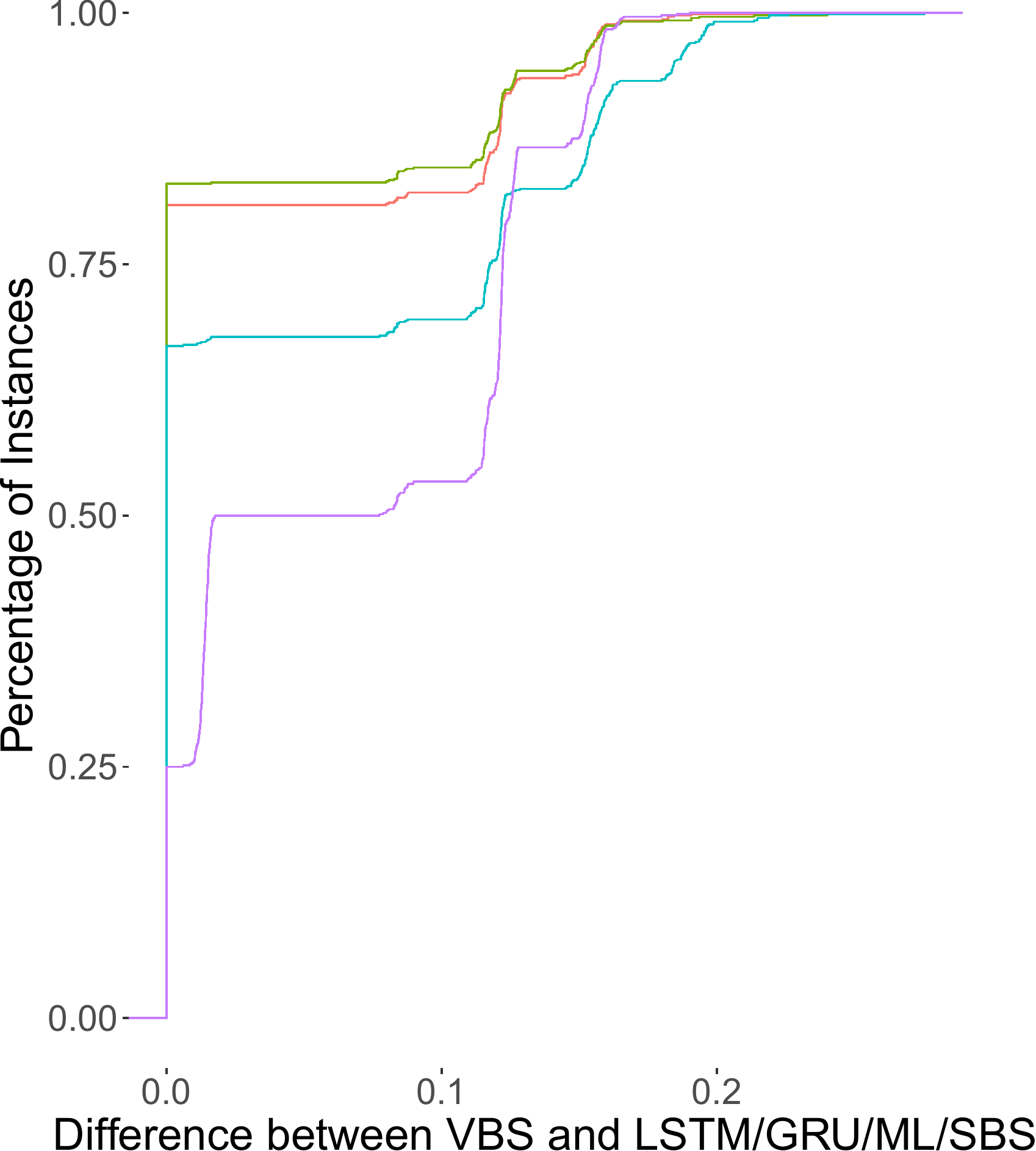}
\subcaption{DS1 - Falkenauer’s Fitness} \label{fig:DS1_Histo_Falk}
\end{minipage}
\begin{minipage}[t]{0.45\linewidth}
\centering
 \includegraphics[angle=0,scale=0.30]{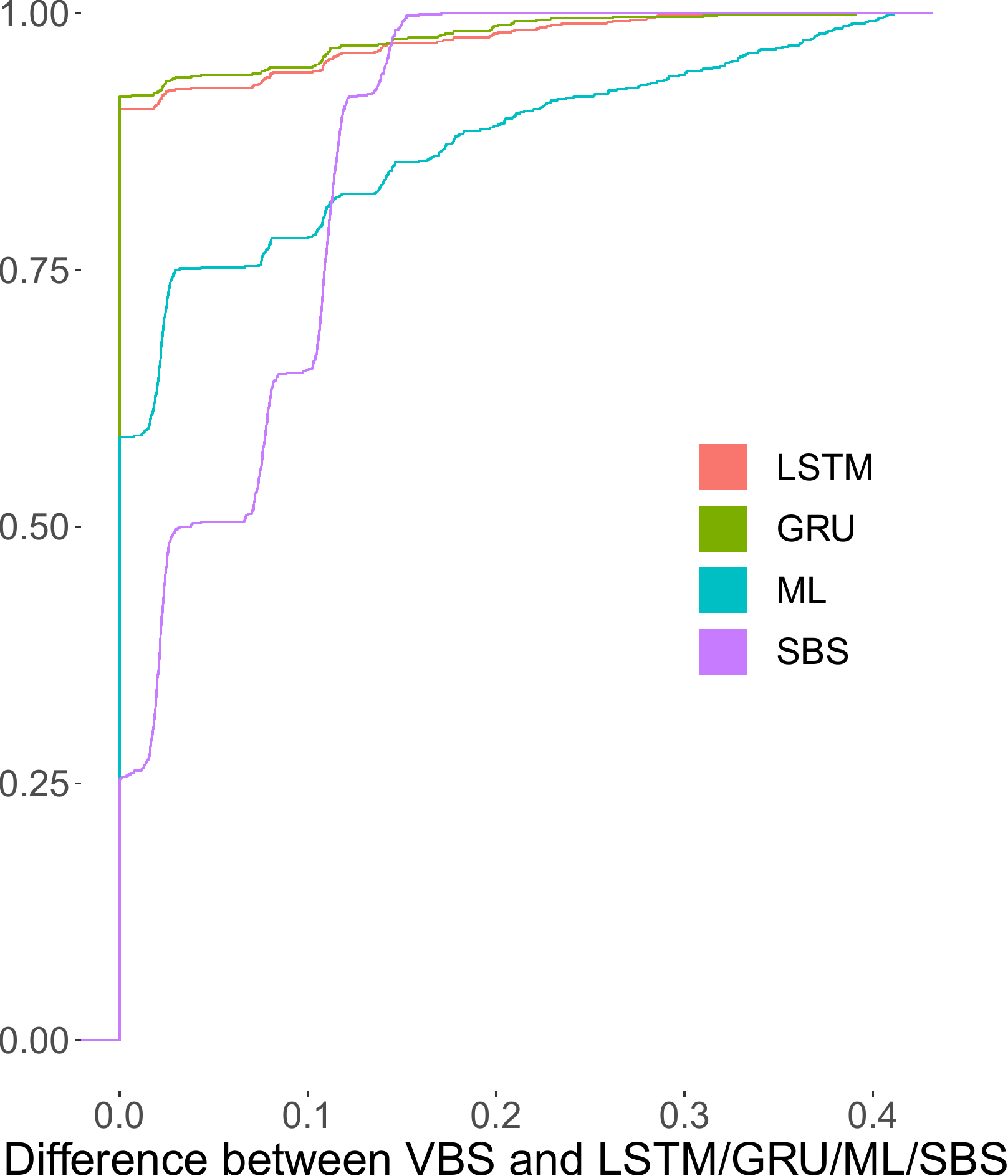}
 \subcaption{DS2 - Falkenauer’s Fitness} \label{fig:DS2_Histo_Falk}
\end{minipage}

\begin{minipage}[t]{0.45\linewidth}
\centering
 \includegraphics[angle=0,scale=0.30]{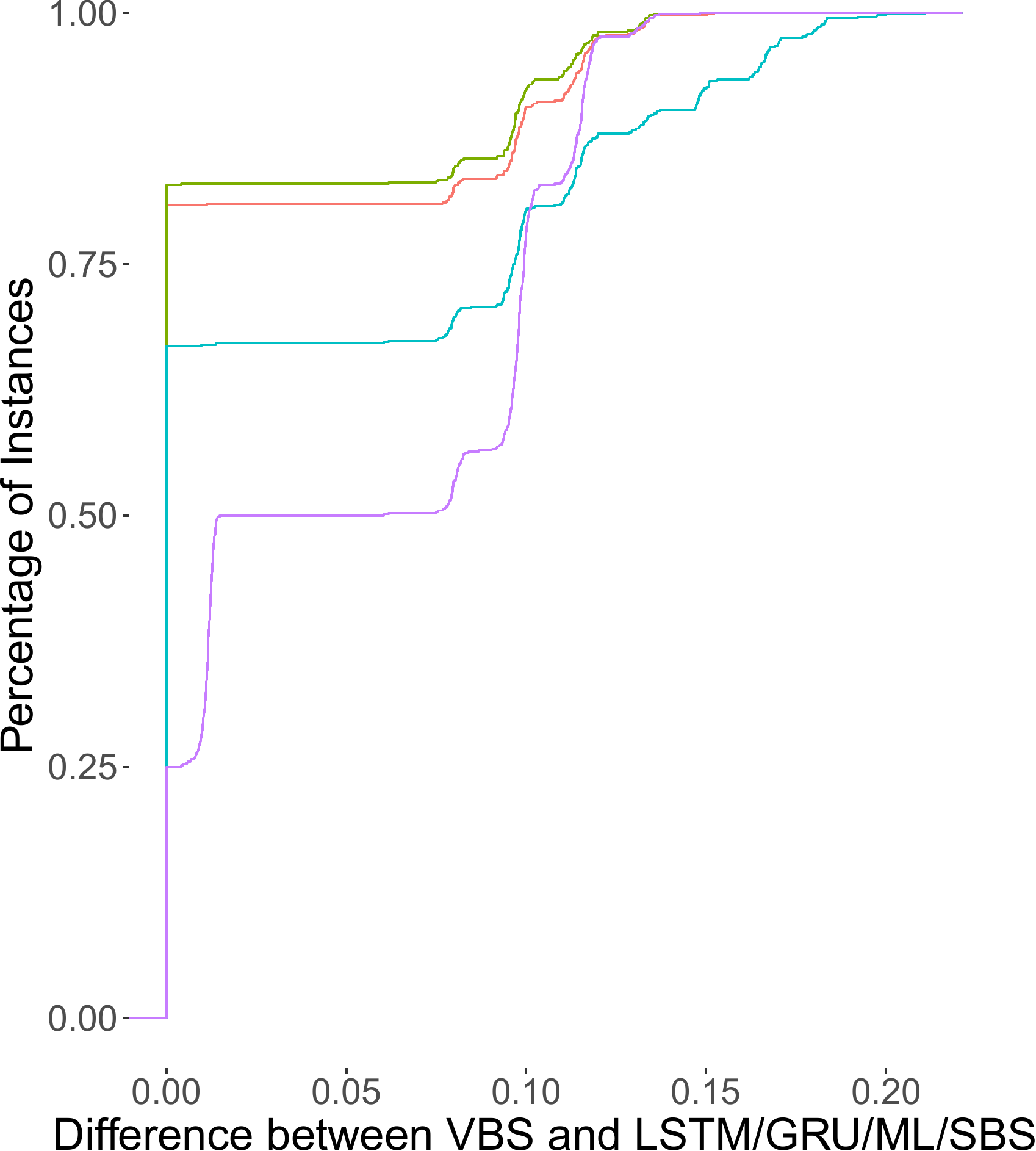}
 \subcaption{DS3 - Falkenauer’s Fitness} \label{fig:DS32_Histo_Falk}
\end{minipage}
\begin{minipage}[t]{0.45\linewidth}
\centering
 \includegraphics[angle=0,scale=0.30]{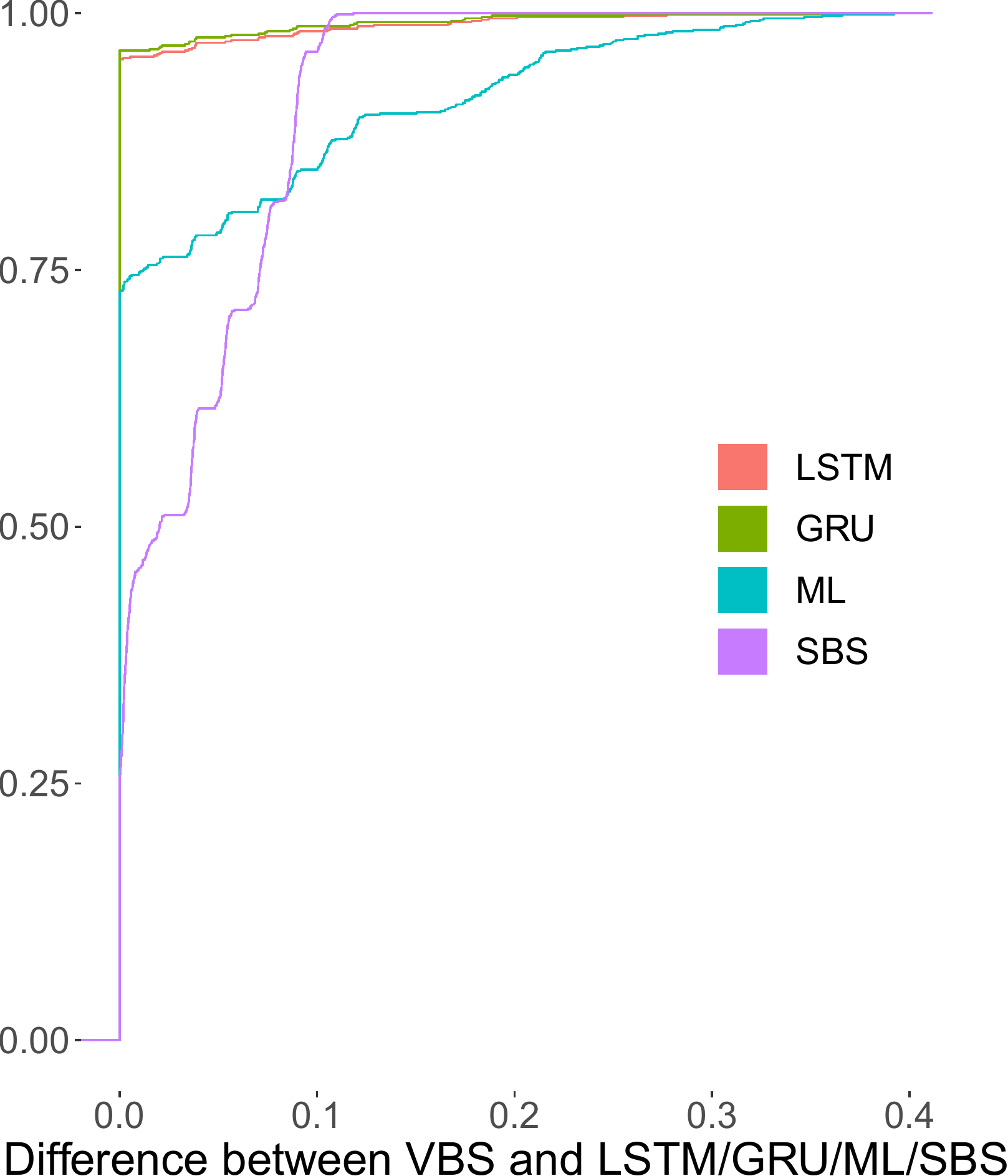}
 \subcaption{DS4 - Falkenauer’s Fitness} \label{fig:DS4_Histo_Falk}
\end{minipage}

\begin{minipage}[t]{0.45\linewidth}
\centering
 \includegraphics[angle=0,scale=0.30]{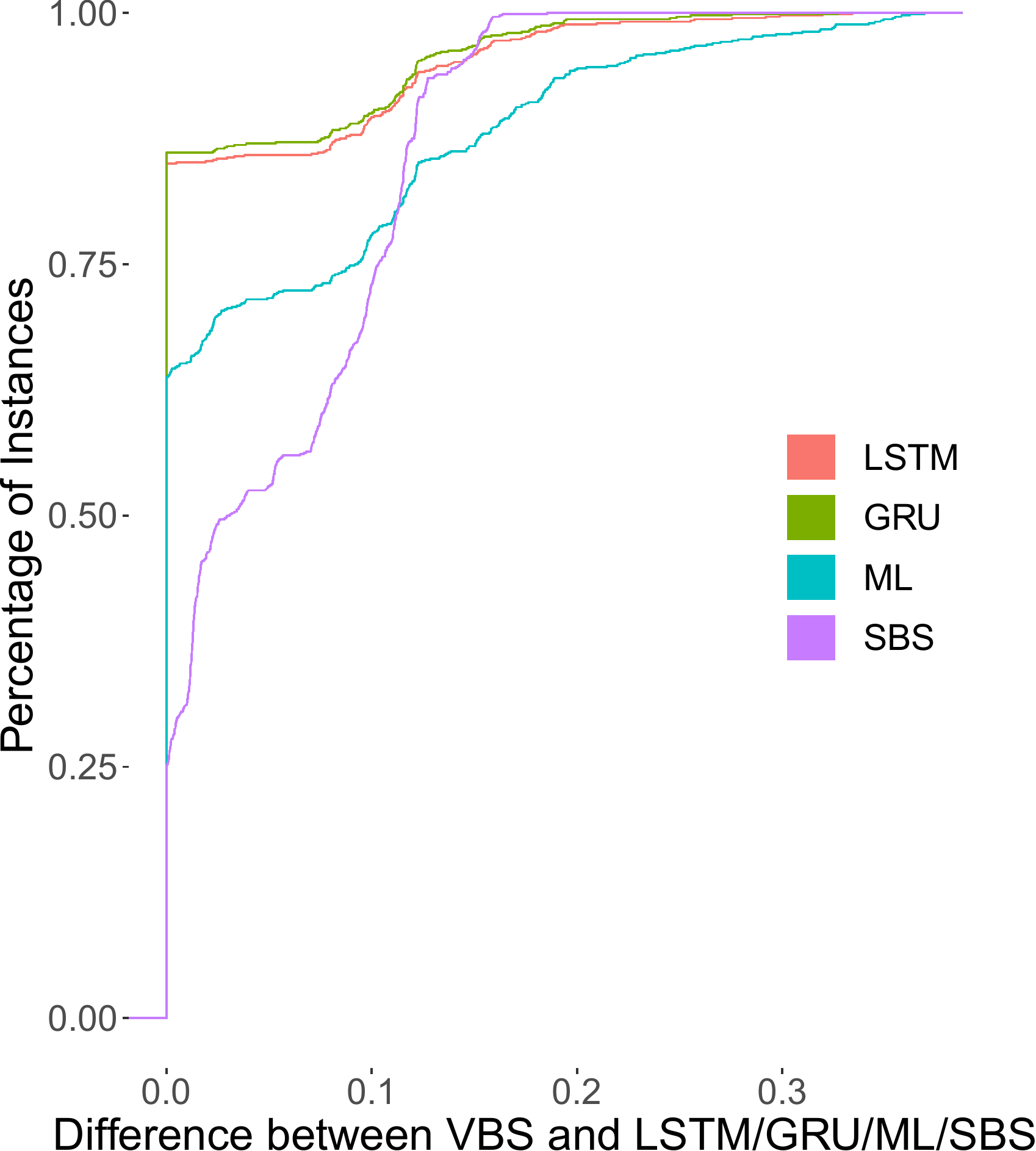}
 \subcaption{DS5 - Falkenauer’s Fitness} \label{fig:DS5_Histo_Falk}
\end{minipage}

\caption{Cumulative distribution plots over the test sets of 800 instances of each DS(1-5) to evaluate LSTM and GRU predictors VS classical ML techniques, SBS and VBS on performance space using Falkenauer's fitness metric}
\label{fig:DS_Histo_Falk}
\end{figure}

\subsection{Evaluation of Solution Quality}

The overall objective of the BPP is to minimise the number of bins used to pack a set of items. Ultimately, the number of containers defines the cost of any real-world solution.
Table \ref{tab:DL_ML_Bins} shows the number of bins required to pack all 800 test instances for each DS(1-5) and contrasts this against the lowest possible number of bins used by the VBS and the number of bins needed using the algorithms predicted by LSTM, GRU and traditional ML selectors.
The LSTM and GRU use between 1.32\% and 2.39\% fewer bins than the SBS and between 0.21\% and 1.52\% more than the VBS. On DS4, GRU uses over 1150 bins fewer than the SBS and only 176 (0.21\%) more than the VBS which uses 83,823 bins. While the ML techniques use between 1.85\% more bins and up to 0.92\% fewer bins than the SBS and 1.87\% to 4.74\% more than the VBS.

After normalising the number of bins $b'$ used by the predicted algorithm (described in section \ref{sec:Methodology}), a Wilcoxon signed-rank test is used to evaluate significance in a pairwise fashion for all comparisons of bins and {\color{black} corrected for multiple comparisons with the Bonferroni method}. Table \ref{tab:Pval-DS[1-5]} in the appendices shows the  p-values of $b'$ in a pairwise fashion to the bins used by the SBS and the VBS. Based on these p-values, Table \ref{tab:ranking_bins} shows the comparison between the LSTM, GRU the best ML technique and the SBS over the different test sets in terms of bins. For all experiments the DL selectors significantly improve on both the SBS (BF for all datasets) and the classical ML techniques (i.e. $\mathcal{H}_0(1,2)$ are rejected), and are very close to the VBS. While the best ML techniques improve over the SBS significantly (i.e. $\mathcal{H}_0(3)$ is rejected) only on DS(1,3).

\begin{figure}[]
\begin{minipage}[t]{0.45\linewidth}
\centering
\includegraphics[angle=0,scale=0.30]{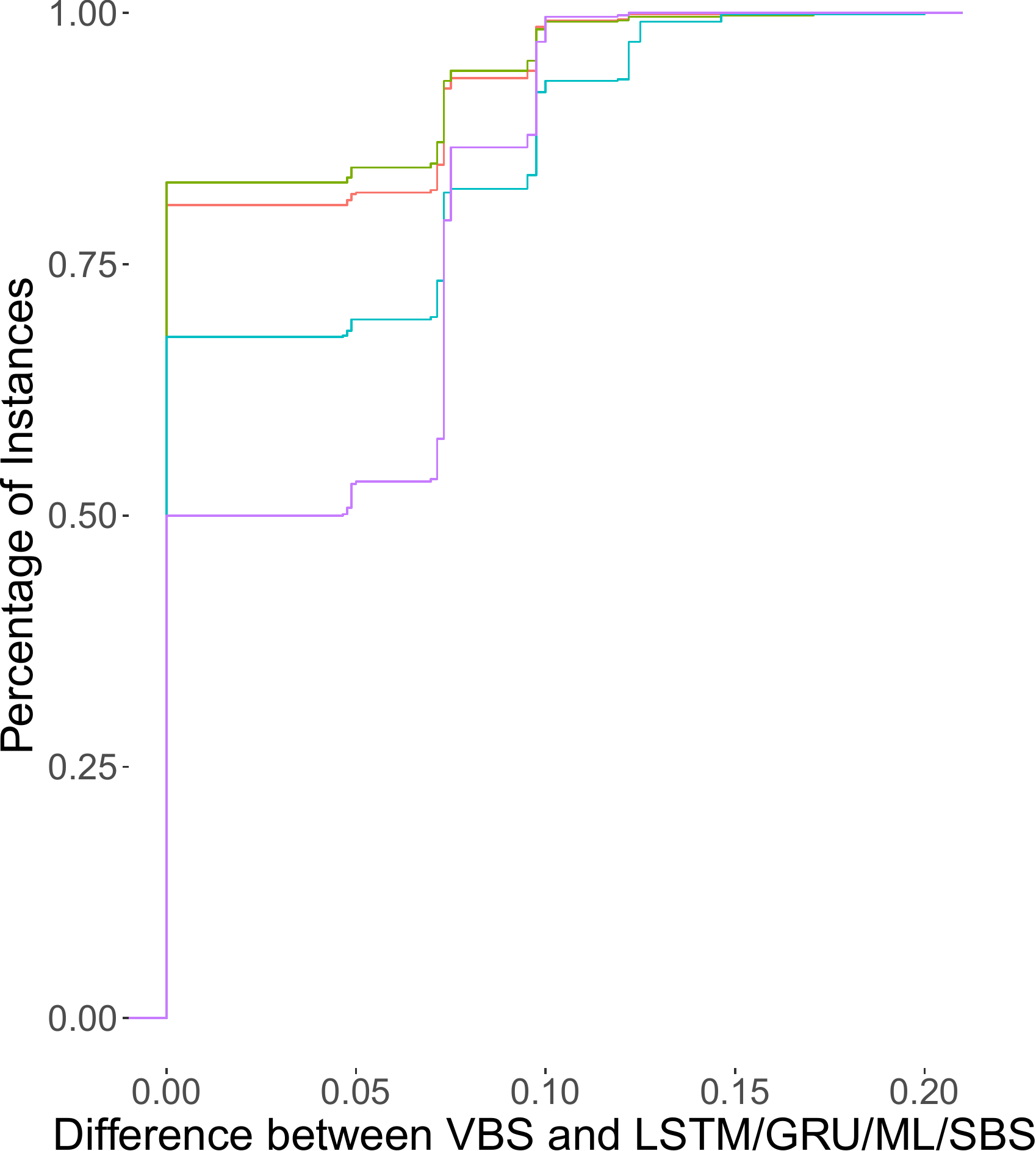}
\subcaption{DS1 - Number of Bins} \label{fig:DS1_Histo_Bins}
\end{minipage}
\begin{minipage}[t]{0.45\linewidth}
\centering
\includegraphics[angle=0,scale=0.30]{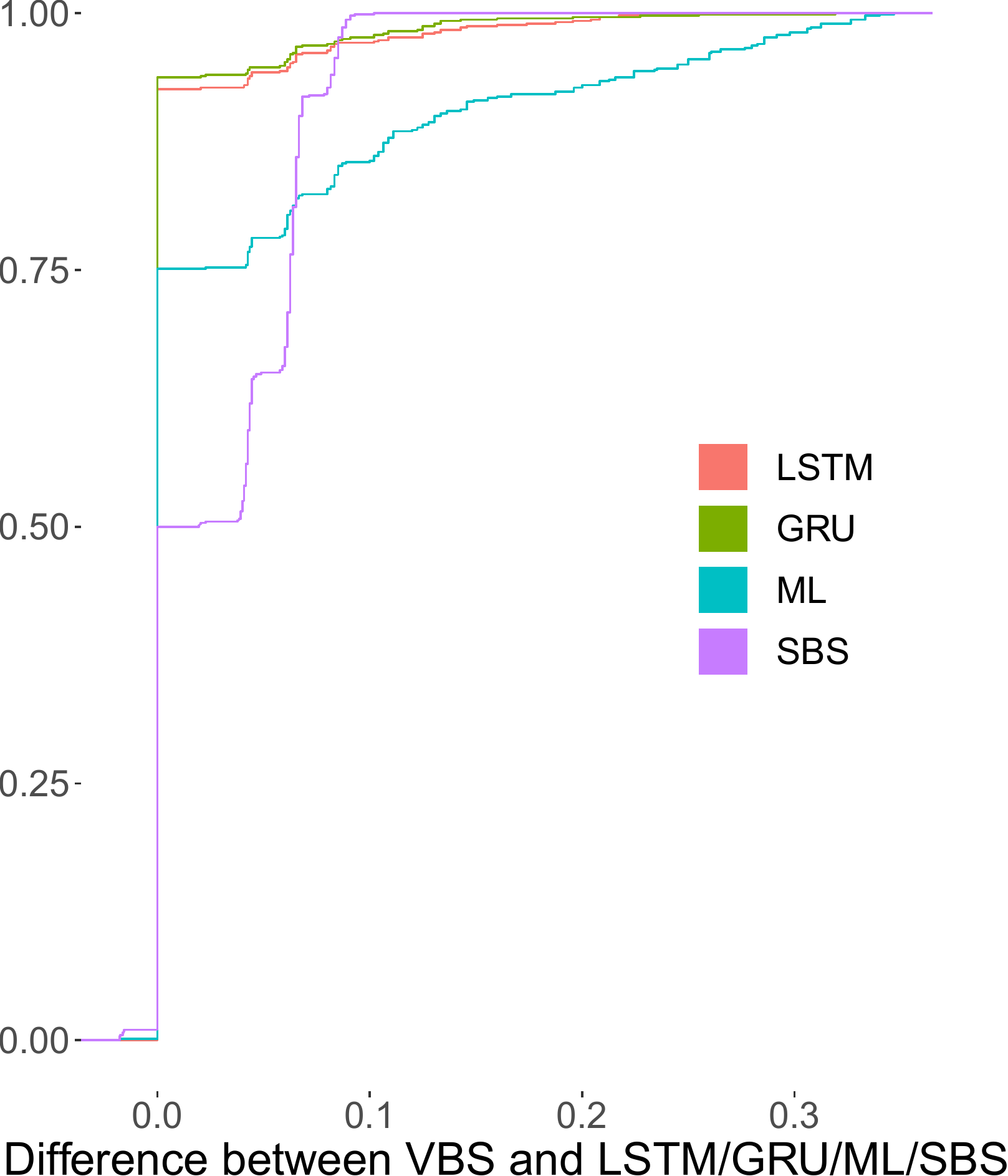}
\subcaption{DS2 - Number of Bins} \label{fig:DS2_Histo_Bins}
\end{minipage}

\begin{minipage}[t]{0.45\linewidth}
\centering
\includegraphics[angle=0,scale=0.30]{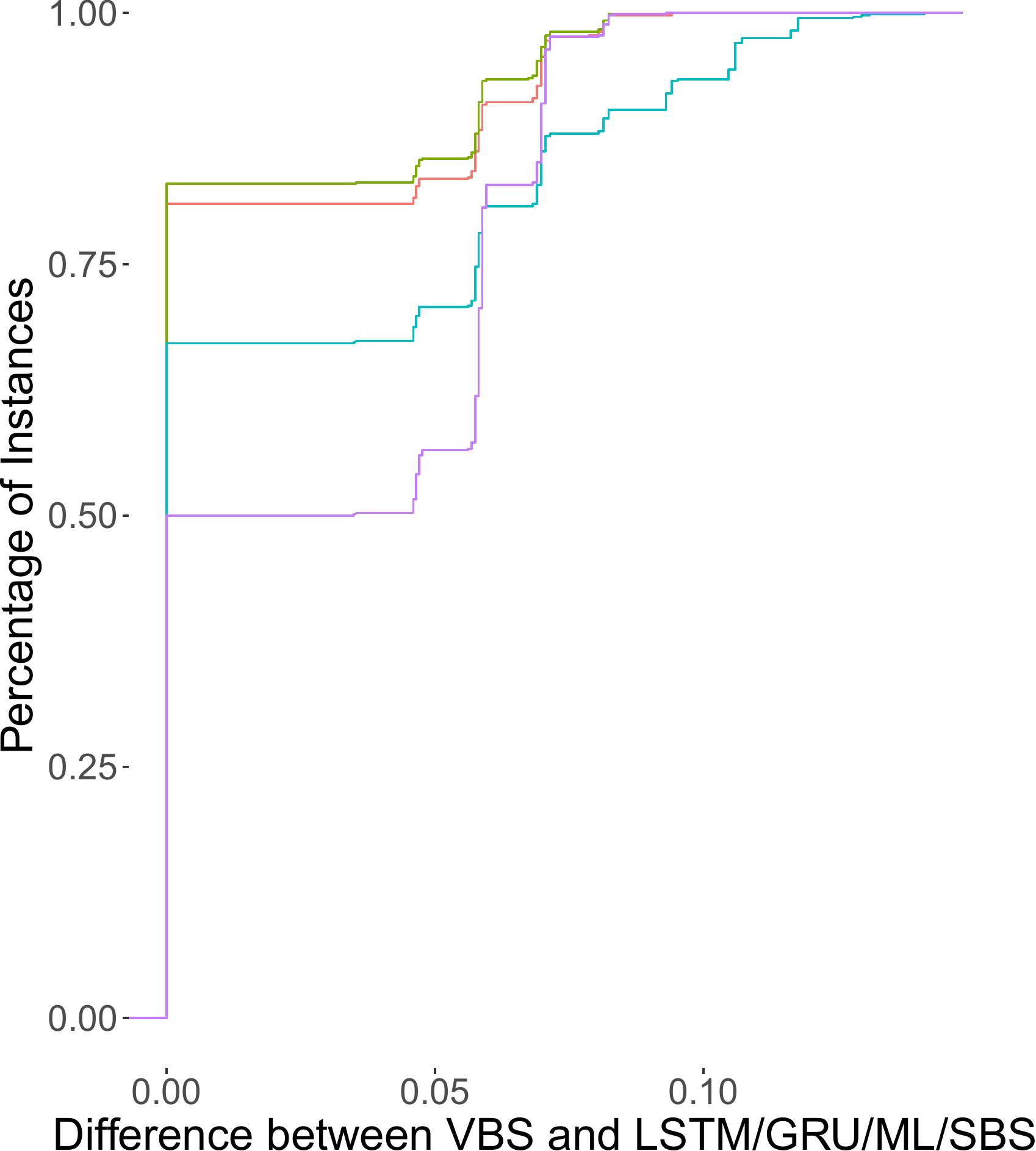}
\subcaption{DS3 - Number of Bins} \label{fig:DS3_Histo_Bins}
\end{minipage}
\begin{minipage}[t]{0.45\linewidth}
\centering
\includegraphics[angle=0,scale=0.30]{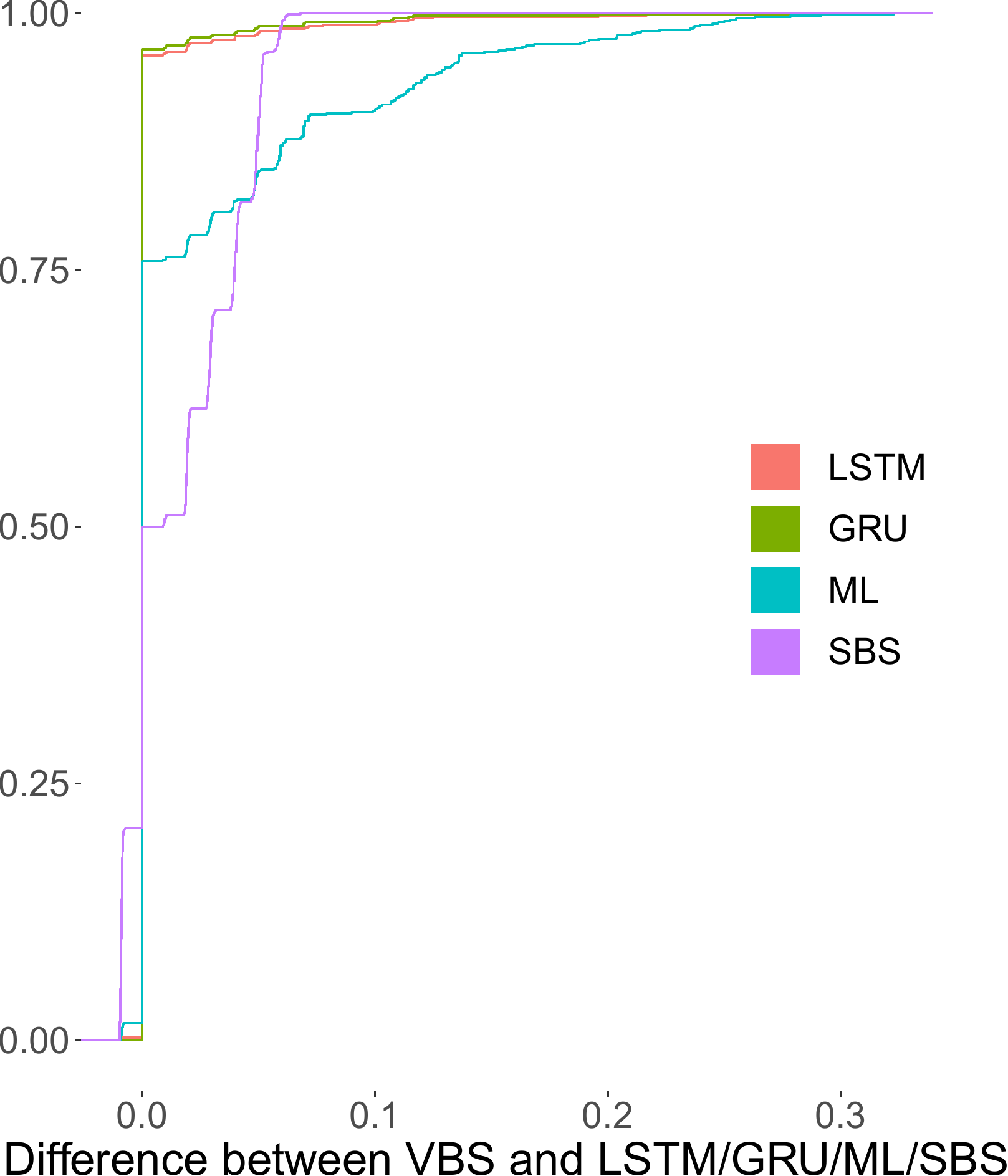}
\subcaption{DS4 - Number of Bins} \label{fig:DS4_Histo_Bins}
\end{minipage}

\begin{minipage}[t]{0.45\linewidth}
\centering
\includegraphics[angle=0,scale=0.30]{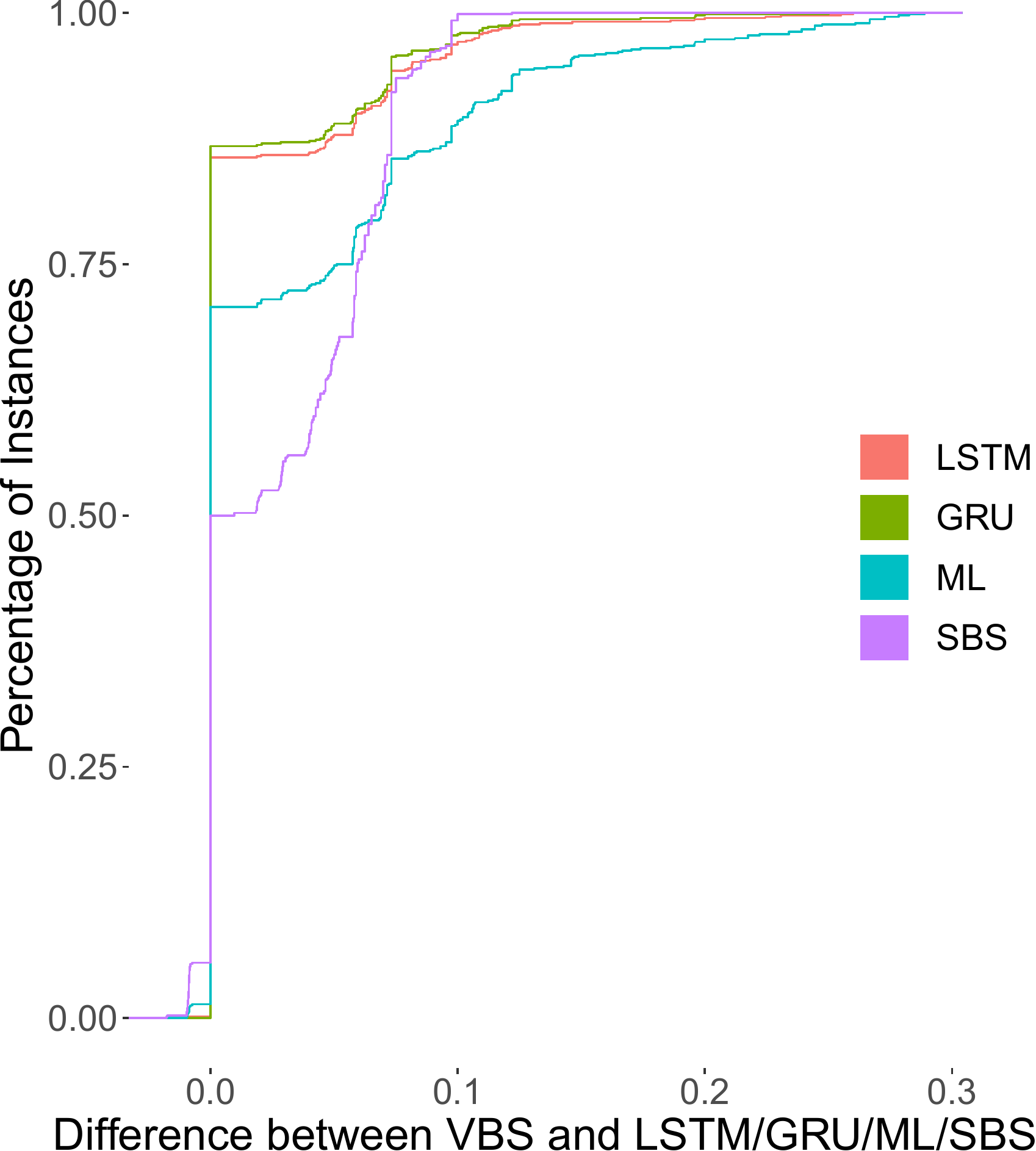}
\subcaption{DS5 - Number of Bins} \label{fig:DS5_Histo_Bins}
\end{minipage}

\caption{Cumulative distribution plots over the test sets of 800 instances of each DS(1-5) to evaluate LSTM and GRU predictors VS classical ML techniques, SBS and VBS on performance space using number of bins metric}
\label{fig:DS_Histo_Bins}
\end{figure}

Figure \ref{fig:DS_Histo_Bins} shows the cumulative distribution plots over the test sets of 800 instances from each DS(1-5) in terms of number of bins, where the difference \textit{d$_{b}$=}(Selector$_{used bins}$ $-$ VBS$_{used bins}$) $\geq$ 0 and \textit{d$_{b}$=0} is the best. In terms of the percentage of instances that are solved by each method, LSTM/GRU solves 82.13\%-98.75\% within 5\% VBS while the best ML technique solves 69.5\%-84.63\% and 53\%-89\% by SBS for DS(1-5). For higher values of differences from the VBS, the results depend on the dataset used. On DS(1,3), 99.13\%-100\% of the instances are solved by LSTM/GRU compared to 93.25\%-93.38\% solved by ML within 10\% of VBS and both techniques solve all the instances within 20\% VBS.
On DS(2,4,5), 99.88\%-100\% instances solved by LSTM/GRU while 98.13\%-100\% by best ML technique within 30\% VBS. The SBS manages to solve 99.6\% to all instances within 10\% VBS on DS(1,3) and within 20\% VBS on DS(2,4,5).

\begin{table}[H]
\caption{Total Bins required to pack instances in the test set for the LSTM predictor, GRU predictor, classical ML techniques, traditional heuristics and VBS, values in \textit{italic} indicate the best ML results and values in \textbf{bold}  the  best overall result per dataset and BF is the SBS over all the datasets.}
\label{tab:DL_ML_Bins}
\resizebox{\columnwidth}{!}{%
\begin{tabular}{ccccc|c|cc|cccccc}
\hline
& \multicolumn{5}{c}{Heuristics} & \multicolumn{2}{c}{DL} & \multicolumn{6}{c}{ML}  \\
\hline
DS & FF & BF(SBS) & WF & NF & VBS & LSTM & GRU & NN & DT & RF & SVM & NB & KNN \\
 \hline
DS1 & 34815 & 34722 & 35357 & 38357 & 33439 & 33947 & \textbf{33891} & \textit{34402} & 34464 & 34414 & 34434 & 34491 & 34423 \\
DS2 & 41494 & 41062 & 43031 & 47961 & 39929 & 40206 & \textbf{40152} & 41234 & 41823 & 41298 & 41322 & \textit{41037} & 41521 \\
DS3 & 71839 & 71741 & 73312 & 79498 & 69638 & 70470 & \textbf{70366} & 71362 & 71465 & 71418 & \textit{71327} & 71680 & 71422 \\
DS4 & 85677 & 85166 & 89435 & 100439 & 83823 & 84043 & \textbf{83999} & 85564 & 86160 & \textit{85492} & 85389 & 85666 & 85806 \\
DS5 & 58536 & 58242 & 60345 & 66672 & 56772 & 57380 & \textbf{57288} & 58592 & 58471 & \textit{58230} & 58615 & 58543 & 58472 \\
\hline
\end{tabular} }
\end{table}

\begin{table}[h]
\caption{The comparison between the LSTM, the GRU, the best ML technique and the SBS over the different test sets in terms of number of bins using the paired Wilcoxon Signed-Rank Test with 5\% confidence level {\color{black} corrected for multiple comparisons with the Bonferroni method}.  For each given pair,  the $\shortuparrow$ means the first approach's median is better, $+$ means there is significance, $\shortdownarrow$ means the first approach's median is worst and $-$ means there is no significance and $\Longleftrightarrow$ means both approaches have same median. }
\centering
\label{tab:ranking_bins}
\begin{tabular}{ccccccc}
\hline
 & \multicolumn{6}{c}{Bins} \\
 \hline
 & LSTM-SBS & LSTM-ML & GRU-SBS & GRU-ML &  ML-SBS & LSTM-GRU \\
 \hline
DS1 &  $\shortuparrow +$ & $\shortuparrow +$ & $\shortuparrow +$ & $\shortuparrow +$ & $\shortuparrow +$ &  $\Longleftrightarrow -$ \\
\hline
DS2 &  $\shortuparrow +$ & $\shortuparrow +$ & $\shortuparrow +$ & $\shortuparrow +$ & $\shortuparrow -$ & $\Longleftrightarrow -$\\
\hline
DS3 & $\shortuparrow +$ & $\shortuparrow +$ & $\shortuparrow +$ & $\shortuparrow +$ & $\shortuparrow +$ & $\Longleftrightarrow -$ \\
\hline
DS4  & $\shortuparrow +$ & $\shortuparrow +$ & $\shortuparrow +$ & $\shortuparrow +$ & $\shortuparrow -$ & $\shortdownarrow -$ \\
\hline
DS5  & $\shortuparrow +$  & $\shortuparrow +$ & $\shortuparrow +$ &  $\shortuparrow +$& $\shortuparrow -$ & $\shortdownarrow -$ \\
\hline
\end{tabular} 
\end{table}

In summary, the results presented indicate that a deep learning method is clearly superior to using a classical prediction method trained with extracted features. {\color{black} Furthermore, as shown in the last column in Tables \ref{tab:ranking_perform} and \ref{tab:ranking_bins}, we infer that the choice of deep method itself has minimal significant effect, i.e. it is the switch to a learning method that captures sequential information that provides the gain in performance.}

\section{A systematic analysis across multiple datasets }
\label{sec:ASP_Rand}
In the previous section we compared the novel feature-free and the traditional feature-based approaches of ASP using datasets in which in the instances were known to be discriminatory with respect to the four heuristics used as solvers. We suggest that the reason that each heuristic favours one subset of instances over another is that each heuristic is able to exploit some implicit structure within the instances in a subset. Thus a model that is able to detect this structure within the instance data can successfully act as an algorithm-selector. In contrast, we suggest that algorithm-selection techniques are likely to perform poorly on instances that do not exhibit exploitable structure.

\begin{figure}[]
\begin{minipage}[t]{0.45\linewidth}
\centering
 \includegraphics[angle=0,scale=0.25]{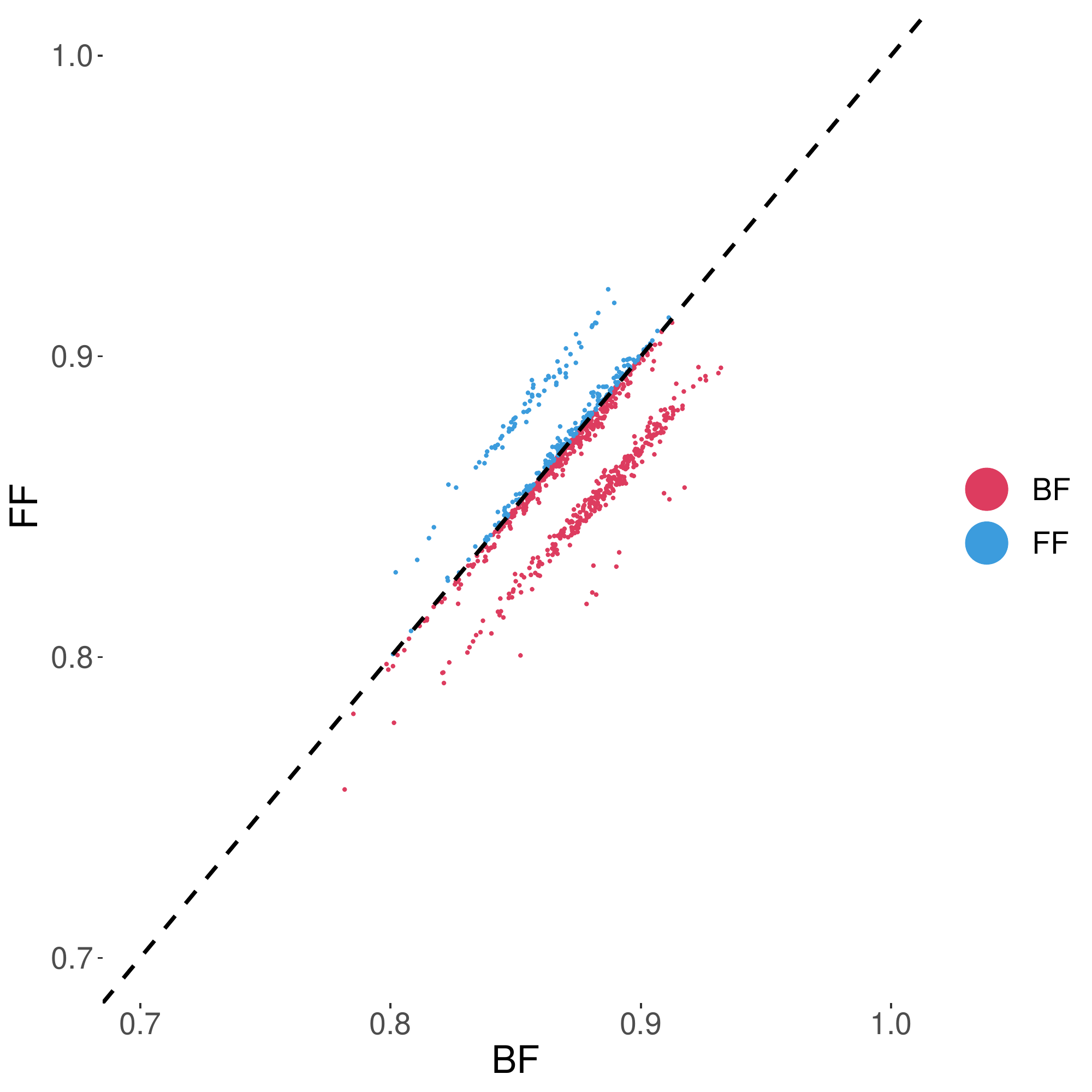}
 \subcaption{Random Instances  Threshold  =  0} \label{fig:DS2_R_Scatter_Falk}
\end{minipage}
\begin{minipage}[t]{0.45\linewidth}
\centering
 \includegraphics[angle=0,scale=0.25]{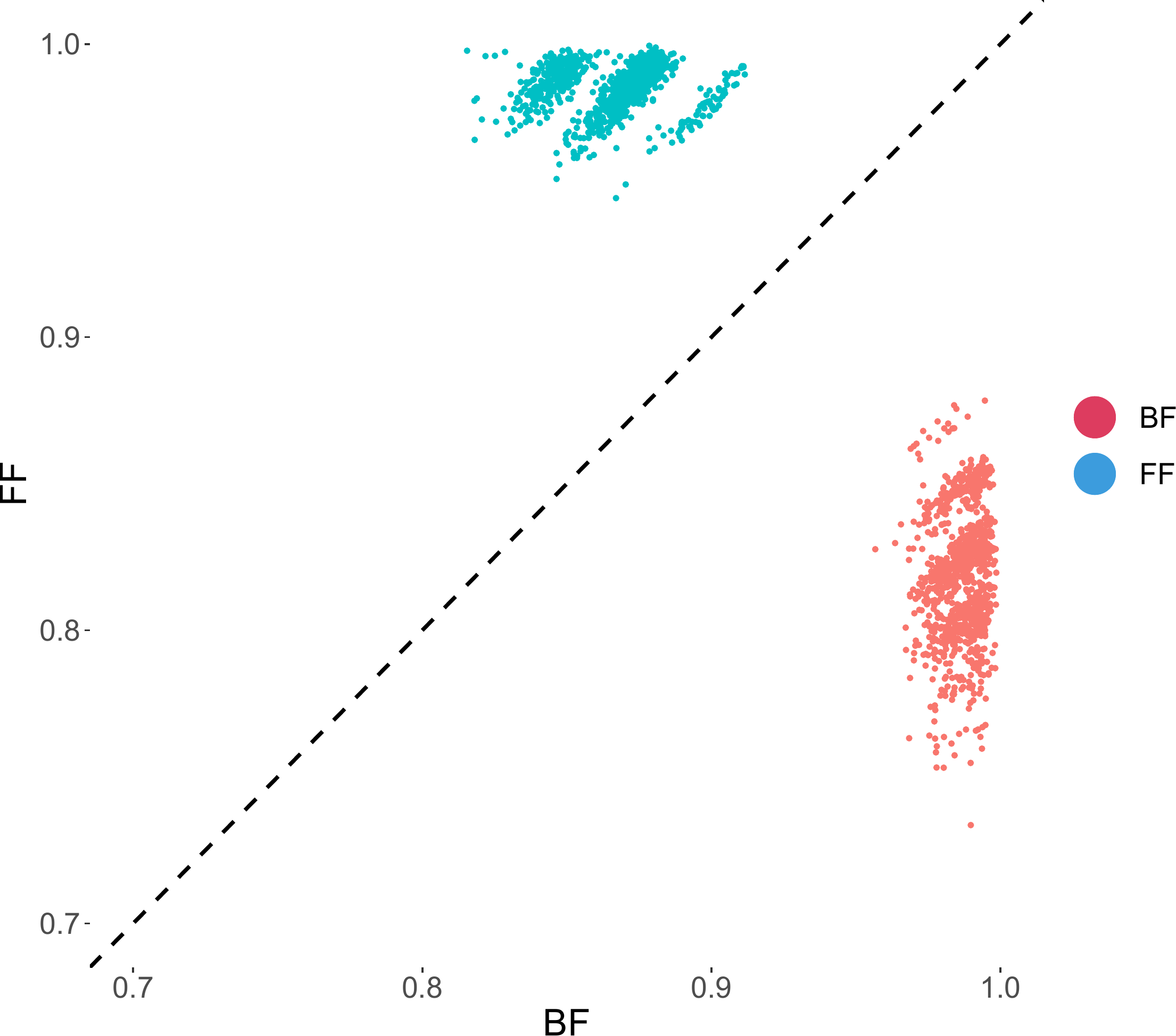}
 \subcaption{Evolved Instances 0.1 $\leq$  Threshold  $\leq$  0.22} \label{fig:DS2_E_Scatter_Falk}
\end{minipage}

\caption{Scatter plot of Falkenauer performance for a) 1000 randomly generated instances and b) 800 evolved instances with 0.1 $\leq$  Threshold  $\leq$  0.22 from the distribution that define datasets DS2 }
\label{fig:1000_RDS2_insts}
\end{figure}

To illustrate this concept, we generate four new datasets: in each, 1000 new instances are generated by selecting item-sizes at randomly from the distributions defined in table \ref{tab:dsGen}. Table \ref{tab:RandInst} shows how many of the randomly generated instances are best solved by each of the 4 heuristics. It is immediately clear that a) The BF heuristic wins the majority of instances in all datasets (73\% of the  instances are best solved by BF);  b) the FF heuristic is the second best heuristic wining approximately the rest of the instances. c) The WF heuristic only wins few instances in two datasets RDS(1,3); d)  the NF heuristic fails to win a single instance in any dataset. The table also highlights that on the same datasets, the BF and FF heuristic tie as winners on large numbers of instances.  Furthermore, even in cases where one heuristic outperforms another, there is very little difference in the performance metric: figure \ref{fig:1000_RDS2_insts} plots the Falkenauer fitness obtained by the two dominant heuristics (BF, FF) on each instances as a scatter-plot. It is immediately clear that most random instances lie on or very close to the diagonal. We suggest therefore that these randomly generated instances contain non exploitable \textit{structure} and hence there is little benefit to be gained from algorithm selection. The same figure-\ref{fig:DS2_E_Scatter_Falk} clearly shows however that the original evolved instances clearly benefit from a selection method.

However, the combinatorial optimisation literature suggests that for many domains,  real-world instances of problems commonly exhibit \textit{structure} that is \textit{not captured by uniform generation of random problems}. Examples of this are described in the TSP domain \cite{hains2011revisiting,hains2012improving} and SAT domain \cite{kroc2009integrating,qasem2009learning} where the authors show that it is important to tailor optimisation methods towards the structured instances in order to obtain good performance rather than relying on `generic' methods. Thus, if algorithm-selector methods are likely to prove beneficial on real-world datasets, this raises a question \textit{`how much structure is required?'}.
In order to shed further light on the relationship between instance structure and algorithm-selection and on the characteristics of datasets on which our proposed technique is likely to work, we conduct a systematic analysis over multiple datasets containing varying levels of structure, as described in the next section.

\begin{table}[h]
\caption{Number of instances are best solved by each heuristics for RDS(1-4)}
\centering
\begin{tabular}{cccccc}
\hline 
 & FF & BF & NF & WF & BF = FF \\
\hline 
RDS1 & 216 & 587 & 0 & 45 & 152 \\
RDS2 & 265 & 735 & 0 & 0 & 0 \\
RDS3 & 327 & 643 & 0 & 10 & 20 \\
RDS4 & 178 & 822 & 0 & 0 & 0 \\
\hline 
\end{tabular}
\label{tab:RandInst}
\end{table}

\subsection{Generating Increasingly Structured Instances}
Based on the analysis above, we consider only BF and FF as potential heuristic solvers. We generate using datasets with that are increasingly discriminatory with respect to these heuristics, using the magnitude of the performance gap between the two heuristics applied to the same instance as a proxy for quantifying implicit structure within the instance that is exploited by one heuristic or the other.  Specifically, we generate instances \textit{at random }from a given distribution, measure the Falkenauer fitness $BF_i$, $FF_i$ on each, then discard instances where $|BF_i-FF_i|< \tau$, for $0 \le \tau \le 0.06$. Thus, higher values of $\tau$ should lead to more structured instances. 1750 instances are generated in this manner from the distributions in table \ref{tab:dsGen} corresponding to DS1 and DS2. In each data set, we accumulate 1000 instances best solved by BF and 750 best solved by FF, to reflect the performance distribution as described in the next section.

\subsection{Training and Evaluation}
From each dataset of 1750 instances, we create a balanced dataset for \textit{training} but evaluate performance on a test-set whose composition is reflective of the distribution of instances. The training set thus contains 500 instances best solved by each BF and FF, while the test set includes 750 instances, 500 instances best solved by BF and 250 for FF to reflect the imbalance performance distribution. 
{\color{black}
Similarly to the previous experiments,  10-fold cross validation is used to train the GRU, NN and RF models taking 10\% of the training instances to verify each fold and to evaluate the generalisation error. Table \ref{tab:GRUValtset} in the appendix shows results achieved on the \textit{validation sets} used during training for the GRU, NN and RF models on RDS$_{\tau}$(1,2). Table \ref{tab:GRUtestset} shows the results achieved on each threshold test set with the models obtained from training on the full training set. We report the classification accuracy as an indicator of the GRU's, NN's and RF's predictive abilities. For RDS$_{\tau}$1, where item-sizes are drawn from a Gaussian distribution, we do not observe any obvious trend as the `structure' in the instances increases for both the feature-free and features-based approaches. As mentioned in Section \ref{sec:Accuracy}, the instances generated using this range of values [40, 60] and Gaussian distribution are more difficult to classify even with the evolved instances (i.e. they have maximum `structure' threshold). In contrast, in RDS$_{\tau}$2 where the item-sizes are drawn from a uniform-random distribution, there is a general trend of feature-free approach performance increasing as $\tau$ increases, with high-level of classification accuracy ($>$75\%) exhibited when $\tau \ge 0.05$. While this trend is much less obvious with the feature-based approaches with classification accuracy at its best is 67.87\%.}

\begin{table}[]
\caption{{\color{black}Classification accuracy of the GRU, NN and ML models in each experiment from the test set for RDS$_{\tau}$(1,2) per threshold. Figures in \textbf{bold} indicate that the model performed significantly better than the SBS in terms of Falkenauer's Performance.} }
\label{tab:GRUtestset}
\begin{tabular}{c|c|ccccccc}
\hline
\multicolumn{2}{c}{} & 0 & 0.01 & 0.02 & 0.03 & 0.04 & 0.05 & 0.06 \\
\hline
\multirow{3}{*}{RDS$_{\tau}$1 } & GRU & 51.60\% & 51.07\% & 53.73\% & 54.13\% & 50.27\% & 52.67\% & 50.93\% \\
 & NN & 53.86\% & 52.93\% & 51.73\% & 57.59\% & 54.40\% & 47.59\% & 44.66\% \\
 & RF & 49.73\% & 51.73\% & 52.53\% & 55.47\% & 53.87\% & 54.00\% & 50.00\% \\
 \hline
\multirow{3}{*}{RDS$_{\tau}$2 } & GRU & 54.27\% & 53.2\% & 52.13\% & 55.47\% & 63.33\% & \textbf{78.13\%} & \textbf{75.46\% } \\
 & NN & 50.53\% & 51.33\% & 61.19\% & 52.66\% & 66.26\% & 58.53\% & 65.06\% \\
 & RF & 52.93\% & 50.80\% & 56.27\% & 54.13\% & 64.80\% & \textbf{67.87\%} & 62.80\% \\
 \hline
\end{tabular} 
\end{table}

\begin{figure}[]
\begin{minipage}[t]{0.45\linewidth}
\centering
\includegraphics[angle=0,scale=0.30]{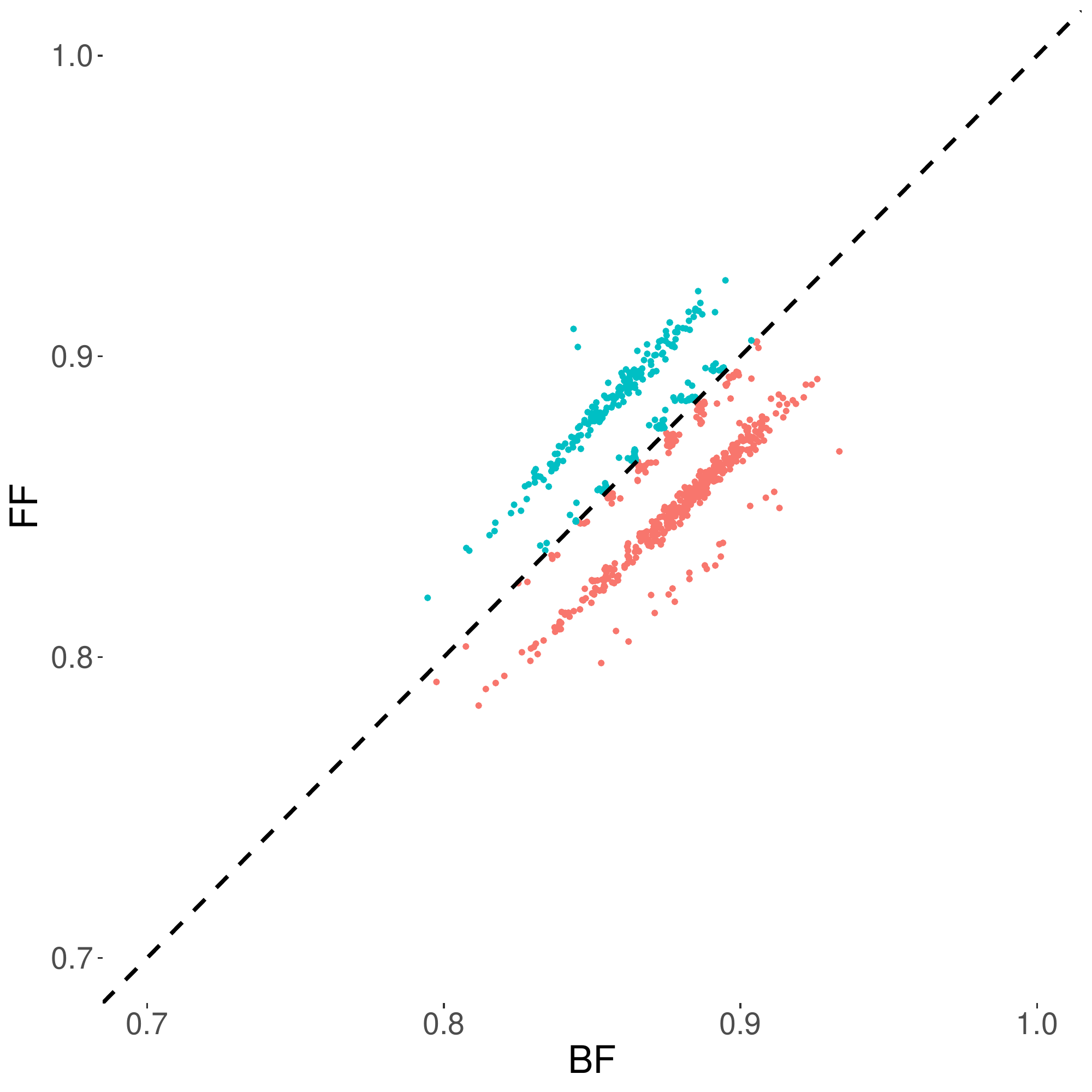}
\subcaption{ Threshold 0} \label{fig:DS2_R_Scatter_Falk_0}
\end{minipage}
\begin{minipage}[t]{0.45\linewidth}
\centering
 \includegraphics[angle=0,scale=0.30]{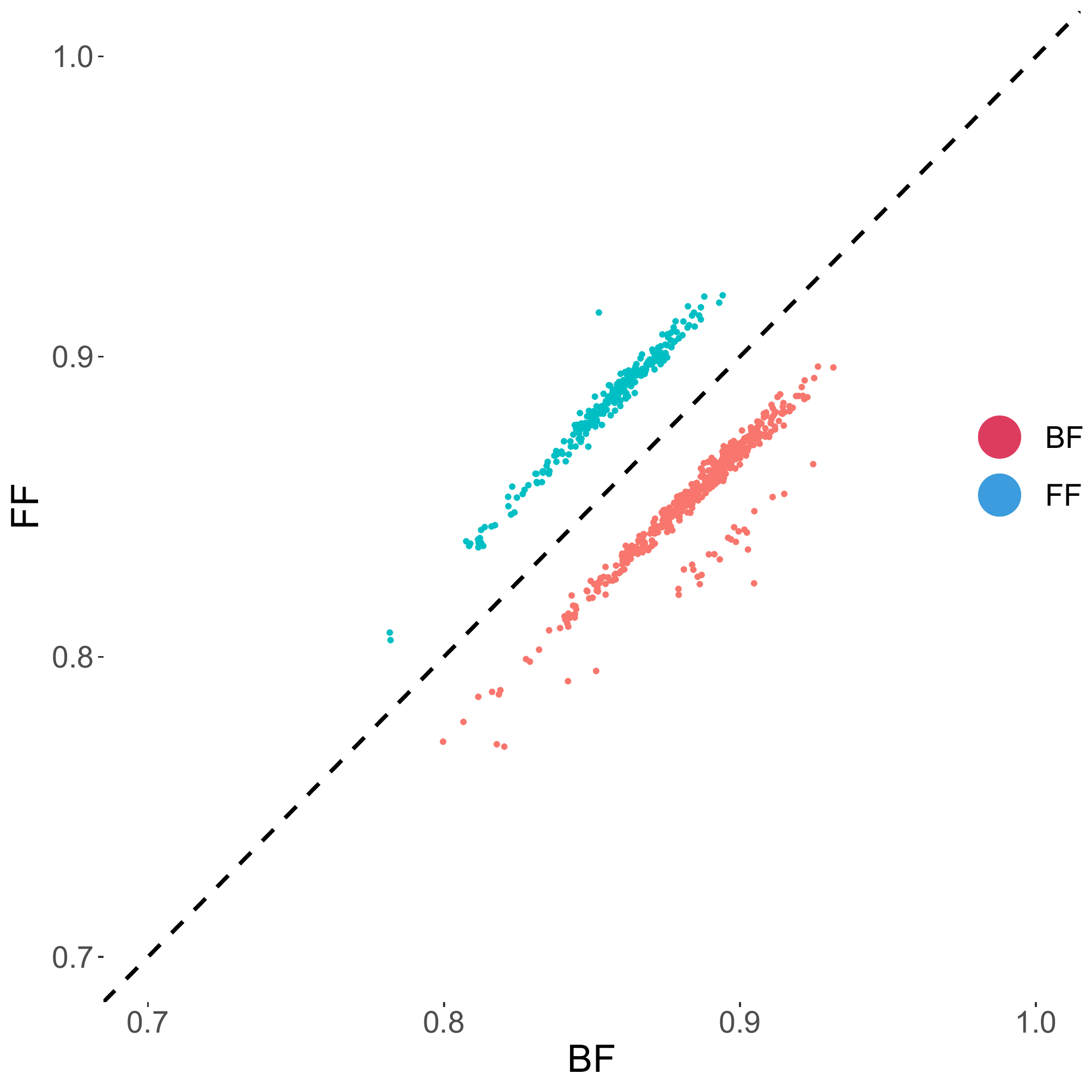}
 \subcaption{ Threshold 0.02} \label{fig:DS2_R_Scatter_Falk_02}
\end{minipage}

\begin{minipage}[t]{0.45\linewidth}
\centering
 \includegraphics[angle=0,scale=0.30]{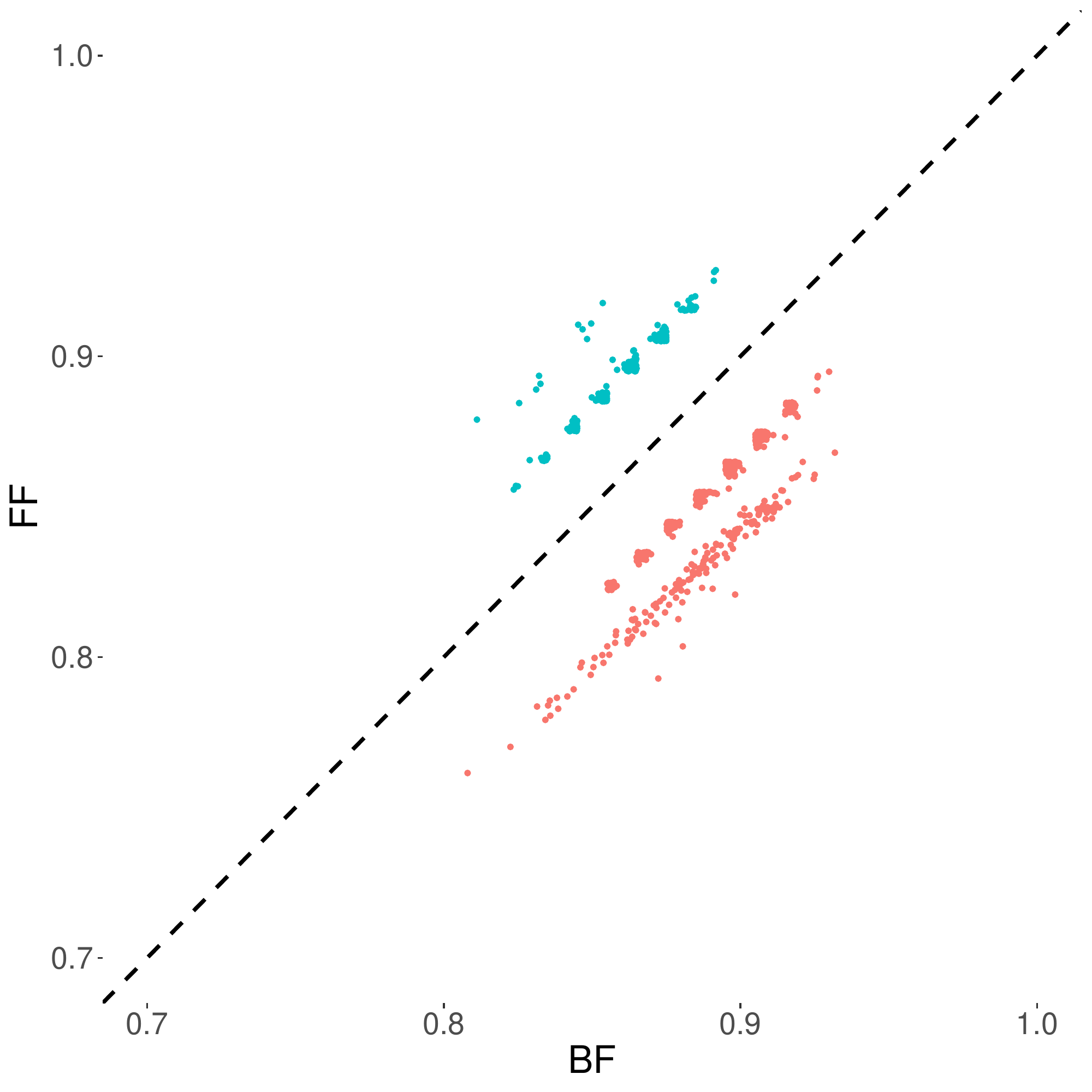}
 \subcaption{Threshold 0.04} \label{fig:DS2_R_Scatter_Falk_04}
\end{minipage}
\begin{minipage}[t]{0.45\linewidth}
\centering
 \includegraphics[angle=0,scale=0.30]{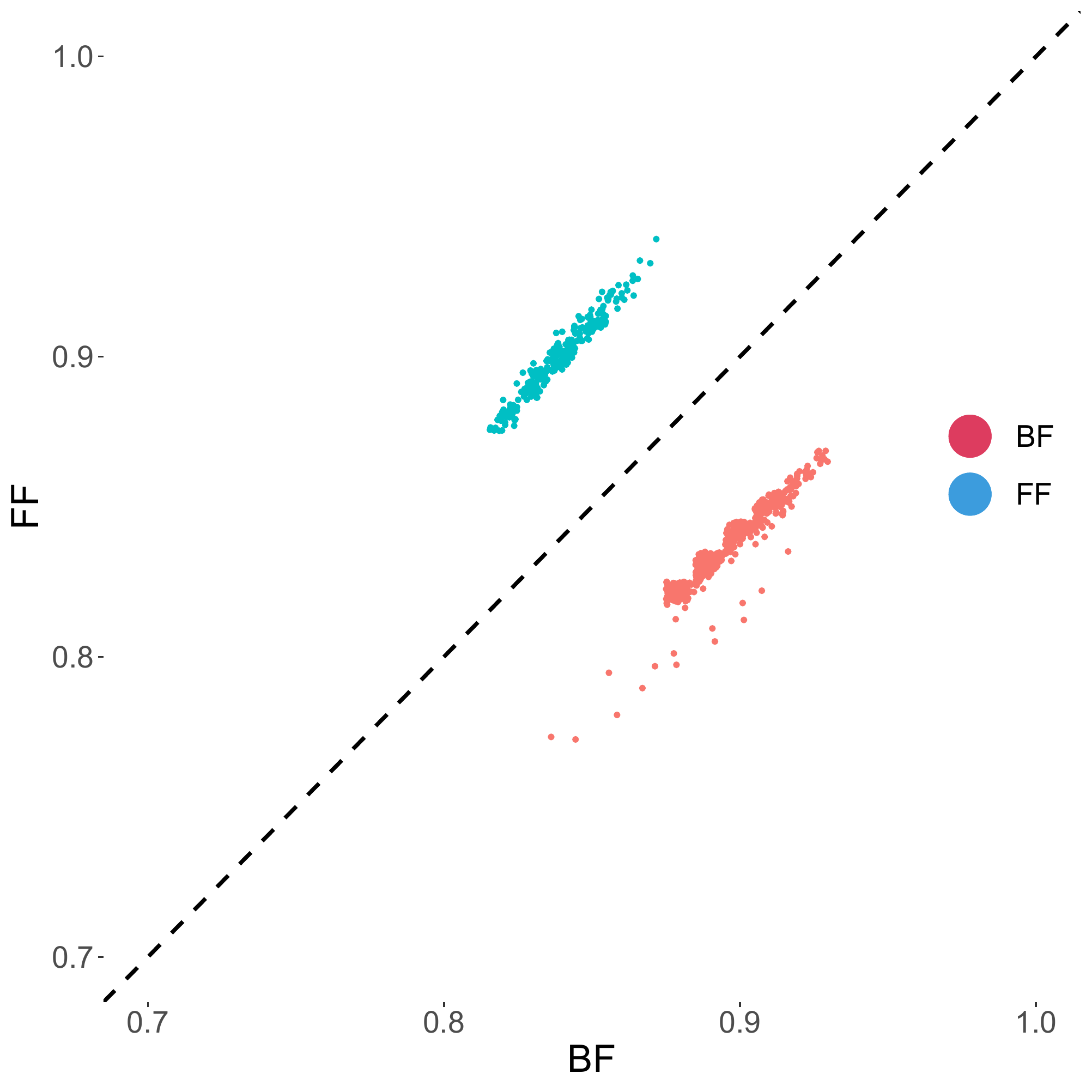}
 \subcaption{ Threshold 0.06} \label{fig:DS2_R_Scatter_Falk_06}
\end{minipage}

\caption{ Scatter Falkenauer’s Fitness  plots with range of different thresholds for test set of RDS$_{\tau}$2 (500 BF and 250 FF).}
\label{fig:DS2_Scatter_Falk}
\end{figure}

{\color{black}
Table \ref{tab:Rand_bins} shows the number of bins required to pack all 750 test instances for each RDS$_{\tau}$(1,2) per threshold and contrasts this against the number of bins used by the SBS and the number of bins needed using the algorithms predicted by each of GRU, NN and RF selectors. In general, the feature-free approach wastes less and saves more bins than the feature-based approaches. On RDS$_{\tau}$1, the GRU uses between 0.18\% and 0.72\% more bins than the SBS, while NN uses between 0.14\% and 1.00\%; RF uses 0.20\% and 0.76\% more bins than the SBS.  Although both approaches fail to save bins with thresholds less than 0.04 on RDS$_{\tau}$2, the GRU uses 0.33\% and 0.43\% fewer bins than the SBS for thresholds 0.06 and 0.05 respectively, NN fails to save any bin and RF saves only 0.04\% using 0.05 threshold. On RDS$_{\tau}$2 with threshold 0.05, GRU uses 170 bins fewer than the SBS while RF saves only 16 bins. Using Wilcoxon signed-rank test, significance is calculated in a pairwise fashion and corrected for multiple comparisons with the Bonferroni method between the GRU, ML techniques and SBS for each threshold of dataset RDS$_{\tau}$(1,2). p-values are shown in table \ref{tab:Pval-RDS(1,2)} in the appendix. Table \ref{tab:Rand_ranking} additionally shows the comparison between the GRU, ML techniques and the SBS over the different thresholds test sets.
}

\begin{table}[]
\caption{{\color{black}Total Bins required to pack instances in the test set for the GRU, NN and RF predictors and the SBS (BF) for RDS$_{\tau}$(1,2) per threshold. Figures in \textbf{bold} indicate the GRU results that are better than the SBS results in terms of the number of bins used.}}
\centering
\label{tab:Rand_bins}
\begin{tabular}{c|c|ccccccc}
\hline
\multicolumn{2}{c}{} & 0 & 0.01 & 0.02 & 0.03 & 0.04 & 0.05 & 0.06 \\
\hline
\multirow{4}{*}{RDS$_{\tau}$1} & BF & 33305 & 33257 & 33229 & 33090 & 32700 & 33037 & 33001 \\
 & GRU & 33366 & 33323 & 33326 & 33184 & 32833 & 33239 & 33237 \\
 & NN & 33352 & 33316 & 33342 & 33157 & 32794 & 33311 & 33331 \\
 & RF & 33373 & 33339 & 33334 & 33173 & 32795 & 33218 & 33251 \\
 \hline
\multirow{4}{*}{RDS$_{\tau}$2} & BF & 38783 & 38814 & 38718 & 38642 & 38228 & 39370 & 38998 \\
 & GRU & 38879 & 38900 & 38837 & 38733 & 38268 & \textbf{39200} & \textbf{38868} \\
 & NN & 38905 & 38907 & 38761 & 38751 & 38245 & 39494 & 39024 \\
 & RF & 38886 & 38917 & 38804 & 38739 & 38257 & \textbf{39354} & 39058 \\
 \hline
\end{tabular}
\end{table}

\begin{table}[]
\caption{{\color{black}The comparison between the feature-free approach using GRU, feature-based approach using ML techniques and the SBS over datasets RDS$_{\tau}$(1,2) per threshold in terms of performance and number of bins using the paired Wilcoxon Signed-Rank Test with 5\% confidence level  corrected for multiple comparisons with the Bonferroni method}. For a given pair of tests, the $\shortuparrow$ means the first approach's median is better, $+$ means there is significance, $\shortdownarrow$ means the first approach's median is worst, $-$ means there is no significance and $\Longleftrightarrow$ means both approaches have same median. }
\label{tab:Rand_ranking}
\resizebox{\columnwidth}{!}{%
\begin{tabular}{c|ccccccc|ccccccc}
\hline
 & \multicolumn{6}{c}{Falkenauer's Performance} & \multicolumn{8}{c}{Bins} \\
 \hline
\textbf{RDS$_{\tau}$1} & 0.00 & 0.01 & 0.02 & 0.03 &  0.04 & 0.05 & 0.06 & 0.00 & 0.01 & 0.02 & 0.03 &  0.04 & 0.05 & 0.06\\
 \hline
GRU-SBS & $\shortdownarrow +$ & $\shortdownarrow +$ & $\shortdownarrow +$ & $\shortdownarrow +$ & $\shortdownarrow +$ & $\shortdownarrow +$ & $\shortdownarrow +$ &  $\Longleftrightarrow +$ & $\Longleftrightarrow +$ & $\Longleftrightarrow +$ & $\Longleftrightarrow +$ & $\shortdownarrow +$ & $\Longleftrightarrow +$ &  $\Longleftrightarrow +$ \\
\hline
NN-SBS & $\shortdownarrow +$ & $\shortdownarrow +$ & $\shortdownarrow +$ & $\shortdownarrow -$ & $\shortdownarrow +$ & $\shortdownarrow +$ & $\shortdownarrow +$ &  $\Longleftrightarrow +$ & $\Longleftrightarrow +$ & $\Longleftrightarrow +$ & $\Longleftrightarrow +$ & $\Longleftrightarrow +$ & $\Longleftrightarrow +$ &  $\shortdownarrow +$ \\
\hline
RF-SBS & $\shortdownarrow +$ & $\shortdownarrow +$ & $\shortdownarrow +$ & $\shortdownarrow +$ & $\shortdownarrow +$ & $\shortdownarrow +$ & $\shortdownarrow +$ &  $\Longleftrightarrow +$ & $\Longleftrightarrow +$ & $\Longleftrightarrow +$ & $\Longleftrightarrow +$ & $\Longleftrightarrow +$ & $\Longleftrightarrow +$ &  $\Longleftrightarrow +$ \\
\hline
GRU-NN & $\shortdownarrow -$ & $\shortdownarrow -$ & $\shortuparrow-$ & $\shortdownarrow -$ & $\shortdownarrow -$ & $\shortuparrow -$ & $\shortuparrow -$ &  $\Longleftrightarrow -$ & $\Longleftrightarrow -$ & $\Longleftrightarrow -$ & $\Longleftrightarrow -$ & $\shortdownarrow -$ & $\Longleftrightarrow -$ &  $\shortuparrow -$ \\
\hline
GRU-RF & $\shortdownarrow -$ & $\shortuparrow -$ & $\shortuparrow -$ & $\shortdownarrow -$ & $\shortdownarrow -$ & $\shortdownarrow -$ & $\shortuparrow -$ &  $\Longleftrightarrow -$ & $\Longleftrightarrow -$ & $\Longleftrightarrow -$ & $\Longleftrightarrow -$ & $\shortdownarrow -$ & $\Longleftrightarrow -$ &  $\Longleftrightarrow -$ \\
\hline
NN-RF & $\shortuparrow -$ & $\shortuparrow -$ & $\shortdownarrow -$ & $\shortuparrow -$ & $\shortuparrow -$ & $\shortdownarrow +$ & $\shortdownarrow -$ &  $\Longleftrightarrow -$ & $\Longleftrightarrow -$ & $\Longleftrightarrow -$ & $\Longleftrightarrow -$ & $\Longleftrightarrow -$ & $\Longleftrightarrow -$ &  $\shortdownarrow -$ \\
\end{tabular} }

\resizebox{\columnwidth}{!}{%
\begin{tabular}{c|ccccccc|ccccccc}
\hline
 & \multicolumn{6}{c}{Falkenauer's Performance} & \multicolumn{8}{c}{Bins} \\
 \hline
\textbf{RDS$_{\tau}$2} & 0.00 & 0.01 & 0.02 & 0.03 &  0.04 & 0.05 & 0.06 & 0.00 & 0.01 & 0.02 & 0.03 &  0.04 & 0.05 & 0.06\\
 \hline
GRU-SBS & $\shortdownarrow +$ & $\shortdownarrow +$ & $\shortdownarrow +$ & $\shortdownarrow +$ & $\shortdownarrow -$ & $\shortuparrow +$ & $\shortuparrow +$ & $\Longleftrightarrow +$ & $\Longleftrightarrow +$ & $\shortdownarrow +$ & $\shortdownarrow +$ &  $\Longleftrightarrow -$ & $\shortuparrow +$ &  $\Longleftrightarrow +$ \\
\hline
NN-SBS & $\shortdownarrow +$ & $\shortdownarrow +$ & $\shortdownarrow +$ & $\shortdownarrow +$ & $\Longleftrightarrow -$ & $\shortdownarrow -$ & $\shortdownarrow -$ & $\Longleftrightarrow +$ & $\Longleftrightarrow +$ & $\shortdownarrow +$ & $\shortdownarrow +$ &  $\Longleftrightarrow -$ & $\Longleftrightarrow +$ &  $\Longleftrightarrow -$ \\
\hline
RF-SBS & $\shortdownarrow +$ & $\shortdownarrow +$ & $\shortdownarrow +$ & $\shortdownarrow +$ & $\shortdownarrow -$ & $\shortuparrow +$ & $\shortdownarrow -$ & $\Longleftrightarrow +$ & $\Longleftrightarrow +$ & $\shortdownarrow +$ & $\shortdownarrow +$ &  $\Longleftrightarrow -$ & $\Longleftrightarrow -$ &  $\Longleftrightarrow -$ \\
\hline
GRU-NN & $\shortuparrow -$ & $\shortuparrow -$ & $\shortdownarrow +$ & $\shortuparrow -$ & $\shortdownarrow -$ & $\shortuparrow +$ & $\shortuparrow +$ & $\Longleftrightarrow -$ & $\Longleftrightarrow -$ & $\Longleftrightarrow +$ & $\Longleftrightarrow -$ &  $\Longleftrightarrow -$ & $\shortuparrow +$ &  $\Longleftrightarrow +$ \\
\hline
GRU-RF & $\shortuparrow -$ & $\shortuparrow -$ & $\shortdownarrow -$ & $\shortuparrow -$ & $\shortdownarrow -$ & $\shortuparrow +$ & $\shortuparrow +$ & $\Longleftrightarrow -$ & $\Longleftrightarrow -$ & $\Longleftrightarrow -$ & $\Longleftrightarrow -$ &  $\Longleftrightarrow -$ & $\shortuparrow +$ &  $\Longleftrightarrow +$ \\
\hline
NN-RF & $\shortdownarrow -$ & $\shortuparrow -$ & $\shortuparrow +$ & $\shortdownarrow -$ & $\shortuparrow -$ & $\shortdownarrow +$ & $\shortuparrow -$ & $\Longleftrightarrow -$ & $\Longleftrightarrow -$ & $\Longleftrightarrow +$ & $\Longleftrightarrow -$ &  $\Longleftrightarrow -$ & $\Longleftrightarrow +$ &  $\Longleftrightarrow -$ \\
\hline
\end{tabular} }
\end{table}

\subsection{ Analysis and discussion  }

{\color{black}
As widely recognised in the literature, developing an understanding of the region(s) of an instance-space in which an algorithm performs well \cite{smith2010understanding} is essential for performing algorithm-selection. This enables training datasets to be collected that include representative instances from the regions of strength of each algorithm considered. The task for the algorithm-selection designer is then to find a mapping between instances and algorithms:  either by extracting informative features from the dataset ('feature-based') or using the raw-instance data as input ('feature-free').

In figure \ref{fig:1000_RDS2_insts} we demonstrated that  two of the best performing heuristics for bin-packing (FF and BF) perform very  similarly on a large set randomly generated instances, i.e. the instances are not discriminatory  with respect to the heuristics we consider: this is line with results that have  previously been reported in multiple domains \citep{cho2008exploiting,Smith-Miles2011} concerning randomly generated benchmarks. This motivated our decision to evolve instances that specifically maximised the performance-gap between heuristics (figure \ref{fig:1000_RDS2_insts}(b)).  We hypothesised that the evolved instances contain some implicit structure that is exploited by a particular heuristic that enables it to perform well.  Furthermore, we suggested that increasing levels of discrimination between algorithms on an instance corresponds to increasing levels of `structure' within the instance-data.  We therefore created fourteen further datasets in which control the level of structure in randomly generated instance using the performance-gap  between two heuristics as a proxy for structure.  We then evaluated the performance of both feature-free and feature-based approaches on these datasets.
}

A systematic analysis showed that in the case of instances generated at uniform-random, increasing the structure dramatically improved the performance of the GRU model in terms of classification accuracy {\color{black} while this is less obvious using the feature-based NN and RF models}. However, there is no discernible trend from instances generated from a Gaussian distribution. One explanation for this is that the combined effect of drawing from a Gaussian distribution and restricting the item-sizes to a narrow range (40-60 in this case) results in instances which are very similar, i.e. in which it is difficult to find unique structure. The number of ties observed between BF and FF on this dataset (see table \ref{tab:RandInst} lends some weight to this).  Alternatively, a different method of creating structure could be considered, for example, in relation to one or more features of the datasets described in section \ref{sec:features}. 

The results reported in section \ref{sec:Results} on DS1-4 and in table \ref{tab:GRUtestset} suggest that the proposed GRU model that uses only the sequence of item-sizes as input performs well on datasets in which some structure exists within the data ($\tau \ge 0.05$). Given that is  well-known that real-world instances are structured, this adds weight to the case for developing good algorithm selectors for structured instances, and for using structured datasets as benchmarks to both compare selection methods and develop new ones. {\color{black}It also suggests that there is likely to be benefit gained from focusing attention on developing new instances that reflect all areas of the instance-space in which an algorithm performs well, and specifically in filling the gaps in an instance-space where we do not currently have data in order to improve selection methods.  For feature-based approaches, once such data is generated, an additional task of developing informative features from this data still remains}.

\section*{Conclusion}

We have described a novel approach to algorithm-selection for sequential optimisation problems that exhibit an ordering with respect to the elements of the problem and how they should be dealt with. Unlike most ASP techniques, the approach does \textit{not} require the design and selection of features to describe an instance. Two deep-neural networks (RNN-LSTSM and RNN-GRU) were trained using the sequence of items representing an instance directly as input to predict the best algorithm to solve the instance. We have compared this feature-free approach with traditional feature-based approaches using ten hand-designed features and six classical ML techniques. Both the novel and the traditional approaches were thoroughly evaluated on 5 different large datasets, exhibiting different numbers of items and different distributions of item-sizes. All classifiers were trained using a large database of instances in which each instance has a distinct best-solver, previously described in \cite{alissa2019algorithm}.

The accuracy of the LSTM and the GRU models ranges from 80-96\% while the ML models ranges from 41\%-72\%, depending on the dataset used. In terms of the percentage of the instances that are solved using DL predictors within a small difference of the VBS Falkenauer's performance ($\leq$ 5\%), we show between 30\%-35\% improvement over the Single Best Solver (SBS), depending on the dataset used. On the other hand, the best ML predictor ranges from 16\%-17\% improvement over the SBS.  As far we aware, this is the first time that such an approach has been used, and represents a significant step forward in algorithm-selection for problems with sequence information, where the difficulties associated with defining suitable features and selecting from large sets of potential features are well understood.

We suggested that the method is able to perform well on datasets that exhibit structure --- a characteristic that is common in many real-world problems. To
understand the extent to which structure plays a role, we developed fourteen new datasets in which the structure in the dataset was gradually increased, despite generating instances at random. We showed that for instances with no structure (according to our proxy measure), algorithm-selection is likely to deliver much benefit, given that several heuristics give identical or very similar performance. {\color{black} On the other hand, our analysis revealed that the DL selector is able to exceed both ML selectors and SBS on random instances generated from a uniform-random distribution using a threshold $\ge $ 5\% difference of Falkenauer performance between BF and FF heuristics.}

It is worth mentioning that our approach can also be applied to domains that do not naturally have sequence information through artificially transforming them into sequences. For example, in  the TSP domain, an instance is defined by a set of coordinates while a solution is a sequence (ordered series of visits). In order to use our approach, a TSP map could be scanned in a fixed pattern providing a sequence of coordinates representing the order of the cities appear on a map, implicitly encapsulating spatial information. Sequences produced in this manner could then be used to train an LSTM or GRU.

Future work will focus on extending the approach to include larger and more complex sets of algorithms with the portfolio to be chosen from such as meta-heuristics, and to applying the method to other domains that have a sequential nature such as flow-shop/job-shop scheduling.  We also intend to investigate if the method can be adapted to \textit{online} problems where continuous streams of items are presented. By using a moving window, only examining the next $n$ items to be packed, our method may be able to adapt to a continuously changing environment. Another line of future work will be to consider hand-crafted features that account for the patterns observed in the sequential data, i.e borrowing ideas from time-series analysis. Ultimately, the goal is to extract knowledge from the trained models in order to gain new insight into the correlation between orderings and predicted results.

\bibliography{bibliography}

\begin{appendices}

\begin{table}[h]
\caption{Range of values that are used in the LSTM and NN hyper-parameters tuning; the table also shows the final selected values}
\label{table:LSTM-NN-Param}
\begin{tabular}{ccccc}
\hline
\textbf{LSTM/GRU} & \textbf{\#Epoch} & \textbf{Batch size} & \textbf{\#Layer} & \textbf{\#Units}  \\
 \hline
Range & [100-700] & [8-128] & [1-4]LSTM/GRU & -  \\
\hline
Best DS(1,2) & 300 & 32 & 2LSTM/GRU (tanh) + FC (softmax) & 32,32,4  \\
\hline
Best DS(3-5) & 700 & 32 & 2LSTM/GRU (tanh)+ FC (softmax)& 32,32,4  \\
\hline
\hline

\multicolumn{5}{c}{\textbf{NN Hyper-parameters}}    \\
\hline
Range & [50-3500] & - & [3-4] & [6-64]\\
\hline
Best & 3000 & 32 & 2(relu) + 1(softmax) & 10,15,4  \\
\hline
 \multicolumn{5}{c}{All the models LSTM, GRU and NN use}\\
  \multicolumn{5}{c}{"adam" Optimizer and "CategoricalCrossentropy" as loss function} \\
 
\hline
\end{tabular}
\end{table}

\begin{table}[h]
\centering
\caption{Range of values over which grid-search was conducted to optimised the hyper-parameters for the ML experiments; the table also shows the final selected values}
\label{tab:MLParam}
\resizebox{\columnwidth}{!}{%
\begin{tabular}{cccccc}
\hline 
\textbf{DT, RF} & \textbf{max\_depth} & \textbf{max\_features} & \textbf{min\_leaf} & \textbf{min\_split} & \textbf{n\_estimators} \\
\hline 
Range & {[}5-100{]} & {[}1-10{]} & {[}2-100{]} & {[}2-100{]} & {[}32-200{]} \\
Best DT & 30 & 5 & 10 & 100 & - \\
Best RF & 50 & 3 & 2 & 50 & 64 \\
\hline 
\textbf{SVM} & \textbf{kernel} & \textbf{gamma} & \textbf{C} & \textbf{degree} & \textbf{decision\_function} \\
\hline 
Sets & \{'linear', 'rbf', 'poly'\} & \{'auto','scale'\} & {[}0.1-1000{]} & {[}0-6{]} & \{'ovo','ovr'\} \\
Best & poly & scale & 1000 & 6 & ovo \\
\hline 
\textbf{KNN} & \textbf{n\_neighbors} & \textbf{weights} & \textbf{algorithm} & \textbf{leaf\_size} & \textbf{P} \\
\hline 
Sets & {[}1-30{]} & \{'uniform', 'distance'\} & auto & {[}30,50,100{]} & {[}1-5{]} \\
Best & 26 & distance & auto & 30 & 1 \\
\hline 
\textbf{NB} & \textbf{var\_smoothing} &  &  &  &  \\
\hline 
Set & {[}1e-09, 1e-01{]} &  &  &   &   \\
Best &1e-09 &   &   &   &  \\
\hline 
\end{tabular} }
\end{table}

\begin{table}[h]
\caption{Classification accuracy and the standard deviation (std) from the validation sets of each ML technique, the LSTM and GRU}
\centering
\resizebox{\columnwidth}{!}{%
\begin{tabular}{ccc|cccccc}
\hline 
 & \multicolumn{2}{c}{DL} &  \multicolumn{6}{c}{ML}  \\
\hline 
Dataset & LSTM & GRU & NN & DT & RF & SVM & NB & KNN  \\
\hline 
DS1 & \begin{tabular}[c]{@{}c@{}} 78.62\% \\  (+/- 4.74\%)\end{tabular} & \begin{tabular}[c]{@{}c@{}} 80.41\% \\ (+/- 2.04\%)\end{tabular}  & \begin{tabular}[c]{@{}c@{}}65.41\% \\(+/- 1.62\%)\end{tabular} & \begin{tabular}[c]{@{}c@{}}62.94\% \\(+/- 1.06\%)\end{tabular} & \begin{tabular}[c]{@{}c@{}}65.78\% \\(+/- 2.02\%)\end{tabular} & \begin{tabular}[c]{@{}c@{}}66.41\% \\(+/- 1.31\%)\end{tabular} & \begin{tabular}[c]{@{}c@{}}61.34\% \\(+/- 2.14\%)\end{tabular} & \begin{tabular}[c]{@{}c@{}}61.75\% \\(+/- 1.13\%)\end{tabular}  \\
\hline 
DS2 & \begin{tabular}[c]{@{}c@{}}89.91\% \\(+/- 1.73\%)\end{tabular} & \begin{tabular}[c]{@{}c@{}} 93.13\% \\(+/- 1.03\%)\end{tabular}  & \begin{tabular}[c]{@{}c@{}}58.66\% \\(+/- 2.02\%)\end{tabular} & \begin{tabular}[c]{@{}c@{}}54.38\% \\(+/- 2.40\%)\end{tabular} & \begin{tabular}[c]{@{}c@{}}58.16\% \\(+/- 2.51\%)\end{tabular} & \begin{tabular}[c]{@{}c@{}}59.00\% \\(+/- 2.69\%)\end{tabular} & \begin{tabular}[c]{@{}c@{}}55.66\% \\(+/- 2.31\%)\end{tabular} & \begin{tabular}[c]{@{}c@{}}54.38\% \\(+/- 2.21\%)\end{tabular}  \\
\hline 
DS3 & \begin{tabular}[c]{@{}c@{}} 80.94\% \\(+/- 3.58\%)\end{tabular} & \begin{tabular}[c]{@{}c@{}} 80.62\% \\(+/- 1.87\%)\end{tabular}  & \begin{tabular}[c]{@{}c@{}}67.12\% \\(+/- 2.03\%)\end{tabular} & \begin{tabular}[c]{@{}c@{}}66.72\% \\(+/- 2.13\%)\end{tabular} & \begin{tabular}[c]{@{}c@{}}67.50\% \\(+/- 1.54\%)\end{tabular} & \begin{tabular}[c]{@{}c@{}}69.00\% \\(+/- 2.53\%)\end{tabular} & \begin{tabular}[c]{@{}c@{}}51.75\% \\(+/- 0.96\%)\end{tabular} & \begin{tabular}[c]{@{}c@{}}64.34\% \\(+/- 1.88\%)\end{tabular}  \\
\hline 
DS4 & \begin{tabular}[c]{@{}c@{}}95.34\% \\(+/- 2.20\%)\end{tabular} & \begin{tabular}[c]{@{}c@{}} 96.62\% \\(+/- 1.22\%)\end{tabular} & \begin{tabular}[c]{@{}c@{}}73.16\% \\(+/- 2.32\%)\end{tabular} & \begin{tabular}[c]{@{}c@{}}69.25\% \\(+/- 2.33\%)\end{tabular} & \begin{tabular}[c]{@{}c@{}}71.94\% \\(+/- 2.80\%)\end{tabular} & \begin{tabular}[c]{@{}c@{}}74.62\% \\(+/- 2.60\%)\end{tabular} & \begin{tabular}[c]{@{}c@{}}66.06\% \\(+/- 3.34\%)\end{tabular} & \begin{tabular}[c]{@{}c@{}}65.12\% \\(+/- 3.20\%)\end{tabular}   \\
\hline 
DS5 & \begin{tabular}[c]{@{}c@{}}82.06\% \\(+/- 1.97\%)\end{tabular} & \begin{tabular}[c]{@{}c@{}} 84.03\% \\(+/- 2.94\%)\end{tabular} & \begin{tabular}[c]{@{}c@{}}62.06\% \\(+/- 2.88\%)\end{tabular} & \begin{tabular}[c]{@{}c@{}}60.47\% \\(+/- 3.25\%)\end{tabular} & \begin{tabular}[c]{@{}c@{}}63.94\% \\(+/- 1.97\%)\end{tabular} & \begin{tabular}[c]{@{}c@{}}57.16\% \\(+/- 2.83\%)\end{tabular} & \begin{tabular}[c]{@{}c@{}}43.00\% \\(+/- 2.80\%)\end{tabular} & \begin{tabular}[c]{@{}c@{}}59.50\% \\(+/- 2.18\%)\end{tabular}   \\
\hline 
\end{tabular} }
\label{tab:MLValtset}
\end{table}

\begin{table}[h]
\caption{P-values between LSTM, GRU, VBS, SBS and ML (best techniques) from the test set (800 instances) in terms of Falkenauer's performance and number of bins achieved by each of the methods mentioned for DS(1-5) using paired Wilcoxon signed-rank test {\color{black} corrected for multiple comparisons with the Bonferroni method}}
\label{tab:Pval-DS[1-5]}
\resizebox{\columnwidth}{!}{%
\begin{tabular}{cccccc|ccccc}
\hline
 & \multicolumn{5}{c}{Falkenauer’s Performance} & \multicolumn{5}{c}{Bins} \\
\hline 
 & DS1 & DS2 & DS3 & DS4 & DS5 & DS1 & DS2 & DS3 & DS4 & DS5 \\
 \hline 
LSTM-VBS & 10$^{-26 }$  & 10$^{-13 }$ & 10$^{-26 }$  & 10$^{-06 }$  & 10$^{-20 }$  & 10$^{-26  }$  & 10$^{-10  }$  & 10$^{-26  }$  & 10$^{-06  }$  & 10$^{-20  }$  \\
\hline 
LSTM-SBS & 10$^{-42 }$  & 10$^{-69 }$ & 10$^{-40 }$  & 10$^{-80 }$  & 10$^{-47 }$  & 10$^{-24  }$  & 10$^{-38  }$  & 10$^{-22  }$  & 10$^{-46  }$  & 10$^{-26  }$  \\
\hline 
LSTM-ML& 10$^{-13 }$  & 10$^{-32 }$ & 10$^{-13 }$  & 10$^{-29 }$  & 10$^{-17 }$  & 10$^{-12  }$  & 10$^{-20  }$  & 10$^{-14  }$  & 10$^{-27  }$  & 10$^{-14  }$  \\
\hline 
GRU-VBS & 10$^{-23 }$  & 10$^{-11 }$ & 10$^{-23 }$  & 10$^{-05 }$  & 10$^{-19 }$  & 10$^{-23  }$  & 10$^{-09  }$  & 10$^{-23  }$  & 10$^{-05  }$  & 10$^{-18  }$  \\
\hline 
GRU-SBS & 10$^{-47 }$  & 10$^{-75 }$ & 10$^{-47 }$  & 10$^{-84 }$  & 10$^{-51 }$  & 10$^{-29  }$  & 10$^{-45  }$  & 10$^{-28  }$  & 10$^{-49  }$  & 10$^{-28  }$  \\
\hline 
GRU-ML & 10$^{-16 }$  & 10$^{-36 }$ & 10$^{-17 }$  & 10$^{-31 }$  & 10$^{-19 }$  & 10$^{-15  }$  & 10$^{-23  }$  & 10$^{-18  }$  & 10$^{-29  }$  & 10$^{-16  }$  \\
\hline 
LSTM-GRU & 1   & 1 &  0.185  &  1  & 1   & 0.956   & 1 & 0.2162  & 1   & 1  \\
\hline 
ML-SBS & 10$^{-16 }$  & 0.0009 & 10$^{-13 }$  & 10$^{-09 }$  & 10$^{-07 }$  & 10$^{-07  }$  & 1  & 10$^{-05  }$  & 1  & 1 \\
\hline 
\end{tabular} }
\end{table}

\begin{table}
\caption{The Confusion Matrix of the LSTM, GRU and Best ML models in experiments on DS(1-4) from the test set}
\centering
\label{tab:DL_ML_CM}
  \begin{subtable}{.57\linewidth}
   \centering
    \scalebox{0.75}{%
    \begin{tabular}{c||cccc|cccc|cccc}
    \hline 
    \multicolumn{1}{c}{ } & \multicolumn{4}{c}{LSTM} & \multicolumn{4}{c}{GRU}  &  \multicolumn{4}{c}{NN} \\
    \hline 
    Heuristic & BF & FF & NF & \multicolumn{1}{c|}{WF} & BF & FF & NF & \multicolumn{1}{c|}{WF} & BF & FF & NF & WF \\
    \hline 
    BF & 132 & 67 & 0 & 1 & 141 & 57 & 0 & 2 & 134 & 33 & 0 & 33 \\
        \hline 
    FF & 64 & 131 & 0 & 5 & 55 & 135 & 0 & 10 & 72 & 71 & 0 & 57 \\
        \hline 
    NF & 0 & 0 & 200 & 0 & 0 & 0 & 199 & 1 & 0 & 2 & 193 & 5 \\
        \hline 
    WF & 4 & 11 & 1 & 184 & 2 & 6 & 3 & 189 & 24 & 37 & 2 & 137 \\
    \hline 
    \end{tabular}    }
     \caption{DS1}
    \label{tab:DL_ML_DS1}
     \end{subtable}%
     
  \begin{subtable}{.57\linewidth}
   \centering
    \scalebox{0.75}{%
    \begin{tabular}{c||cccc|cccc|cccc}
    \hline 
    \multicolumn{1}{c}{ } & \multicolumn{4}{c}{LSTM} & \multicolumn{4}{c}{GRU}  &  \multicolumn{4}{c}{RF} \\
    \hline 
    Heuristic & BF & FF & NF & \multicolumn{1}{c|}{WF} & BF & FF & NF & \multicolumn{1}{c|}{WF} & BF & FF & NF & WF \\
    \hline 
    BF & 192 & 7 & 0 & 1 & 193 & 6 & 1 & 0 & 162 & 23 & 9 & 6 \\
    \hline 
    FF & 10 & 179 & 0 & 11 & 12 & 175 & 0 & 13 & 24 & 110 & 24 & 42 \\
    \hline 
    NF & 0 & 2 & 184 & 14 & 0 & 4 & 185 & 11 & 23 & 39 & 68 & 70 \\
    \hline 
    WF & 2 & 17 & 11 & 170 & 1 & 12 & 5 & 182 & 2 & 40 & 29 & 129 \\
    \hline 
    \end{tabular}    }
    \caption{DS2}
    \label{tab:DL_ML_DS2}
     \end{subtable}%

  \begin{subtable}{.57\linewidth}
   \centering
        \scalebox{0.75}{%
    \begin{tabular}{c||cccc|cccc|cccc}
    \hline 
    \multicolumn{1}{c}{ } & \multicolumn{4}{c}{LSTM} & \multicolumn{4}{c}{GRU}  &  \multicolumn{4}{c}{SVM} \\
    \hline 
    Heuristic & BF & FF & NF & \multicolumn{1}{c|}{WF} & BF & FF & NF & \multicolumn{1}{c|}{WF} & BF & FF & NF & WF \\
    \hline 
    BF & 123 & 77 & 0 & 0 & 132 & 68 & 0 & 0 & 116 & 55 & 0 & 29 \\
    \hline 
    FF & 72 & 127 & 0 & 1 & 65 & 134 & 0 & 1 & 69 & 75 & 0 & 56 \\
    \hline 
    NF & 0 & 0 & 199 & 1 & 0 & 1 & 199 & 0 & 0 & 0 & 198 & 2 \\
    \hline 
    WF & 1 & 1 & 0 & 198 & 0 & 2 & 0 & 198 & 19 & 35 & 0 & 146\\
        \hline 
        \end{tabular}}
    \caption{DS3}
    \label{tab:DL_ML_DS3}
     \end{subtable}%

  \begin{subtable}{.57\linewidth}
   \centering
        \scalebox{0.75}{%
    \begin{tabular}{c||cccc|cccc|cccc}
    \hline 
    \multicolumn{1}{c}{ } & \multicolumn{4}{c}{LSTM} & \multicolumn{4}{c}{GRU}  &  \multicolumn{4}{c}{NN} \\
    \hline 
    Heuristic & BF & FF & NF & \multicolumn{1}{c|}{WF} & BF & FF & NF & \multicolumn{1}{c|}{WF} & BF & FF & NF & WF \\
    \hline 
    BF & 192 & 7 & 0 & 1 & 195 & 5 & 0 & 0 & 149 & 40 & 0 & 11 \\
    \hline 
    FF & 4 & 190 & 1 & 5 & 5 & 189 & 1 & 5 & 33 & 122 & 6 & 39 \\
    \hline 
    NF & 0 & 0 & 197 & 3 & 0 & 0 & 199 & 1 & 0 & 9 & 176 & 15 \\
    \hline 
    WF & 2 & 11 & 2 & 185 & 0 & 11 & 1 & 188 & 8 & 35 & 21 & 136\\ 
    \hline 
    \end{tabular}}
    \caption{DS4}
  \label{tab:DL_ML_DS4}
     \end{subtable}%

\end{table}

\begin{table}[]
\caption{{\color{black}Classification accuracy and the standard deviation (std) from the validation sets of the GRU, RF and NN for RDS$_{\tau}$(1,2) per threshold }}
\centering
\resizebox{\columnwidth}{!}{%
\begin{tabular}{c|ccccccc}
\hline
 & 0 & 0.01 & 0.02 & 0.03 & 0.04 & 0.05 & 0.06 \\
 \hline
 \multicolumn{8}{c}{\textbf{RDS$_{\tau}$1}} \\
  \hline
 GRU & \begin{tabular}[c]{@{}c@{}}55.70\% \\ (+/- 3.03\%)\end{tabular} & \begin{tabular}[c]{@{}c@{}}52.40\% \\ (+/- 3.32\%)\end{tabular} & \begin{tabular}[c]{@{}c@{}}52.60\% \\ (+/- 3.50\%)\end{tabular} & \begin{tabular}[c]{@{}c@{}}56.40\% \\ (+/- 5.61\%)\end{tabular} & \begin{tabular}[c]{@{}c@{}}49.00\% \\ (+/- 4.29\%)\end{tabular} & \begin{tabular}[c]{@{}c@{}}52.85\%\\  (+/- 2.87\%)\end{tabular} & \begin{tabular}[c]{@{}c@{}}53.50\% \\ (+/- 6.41\%) \end{tabular}  \\
\hline

 NN & \begin{tabular}[c]{@{}c@{}}51.90\% \\ (+/- 4.48\%)\end{tabular} & \begin{tabular}[c]{@{}c@{}}49.10\% \\ (+/- 6.36\%)\end{tabular} & \begin{tabular}[c]{@{}c@{}}54.30\% \\ (+/- 4.92\%)\end{tabular} & \begin{tabular}[c]{@{}c@{}}55.00\% \\ (+/- 4.56\%)\end{tabular} & \begin{tabular}[c]{@{}c@{}}50.90\% \\ (+/- 4.39\%)\end{tabular} & \begin{tabular}[c]{@{}c@{}}52.70\%\\  (+/- 4.23\%)\end{tabular} & \begin{tabular}[c]{@{}c@{}}51.80\% \\ (+/- 2.71\%) \end{tabular}  \\
\hline

 RF & \begin{tabular}[c]{@{}c@{}}47.80\% \\ (+/- 4.38\%)\end{tabular} & \begin{tabular}[c]{@{}c@{}}50.10\% \\ (+/- 3.30\%)\end{tabular} & \begin{tabular}[c]{@{}c@{}}52.70\% \\ (+/- 2.83\%)\end{tabular} & \begin{tabular}[c]{@{}c@{}}55.10\% \\ (+/- 4.25\%)\end{tabular} & \begin{tabular}[c]{@{}c@{}}50.20\% \\ (+/- 3.37\%)\end{tabular} & \begin{tabular}[c]{@{}c@{}}53.10\%\\  (+/- 3.87\%)\end{tabular} & \begin{tabular}[c]{@{}c@{}}47.80\% \\ (+/- 5.31\%) \end{tabular}  \\
\hline
 \multicolumn{8}{c}{\textbf{RDS$_{\tau}$2}} \\

\hline
 GRU &\begin{tabular}[c]{@{}c@{}}52.80\% \\ (+/- 4.62\%)\end{tabular} & \begin{tabular}[c]{@{}c@{}}53.40\% \\ (+/- 4.41\%)\end{tabular} & \begin{tabular}[c]{@{}c@{}}52.90\% \\ (+/- 3.88\%)\end{tabular} & \begin{tabular}[c]{@{}c@{}}51.80\% \\ (+/- 4.19\%)\end{tabular} & \begin{tabular}[c]{@{}c@{}}60.10\% \\ (+/- 5.07\%)\end{tabular} & \begin{tabular}[c]{@{}c@{}}76.00\% \\ (+/- 3.55\%)\end{tabular} & \begin{tabular}[c]{@{}c@{}}71.30\% \\ (+/- 4.94\%)\end{tabular} \\
\hline

 NN &\begin{tabular}[c]{@{}c@{}}53.70\% \\ (+/-  3.90\%)\end{tabular} & \begin{tabular}[c]{@{}c@{}}57.00\% \\ (+/- 8.38\%)\end{tabular} & \begin{tabular}[c]{@{}c@{}}59.70\% \\ (+/- 3.74\%)\end{tabular} & \begin{tabular}[c]{@{}c@{}}53.90\% \\ (+/-  3.21\%)\end{tabular} & \begin{tabular}[c]{@{}c@{}}61.40\% \\ (+/- 5.50\%)\end{tabular} & \begin{tabular}[c]{@{}c@{}}67.80\% \\ (+/- 3.40\%)\end{tabular} & \begin{tabular}[c]{@{}c@{}}65.30\% \\ (+/- 4.41\%)\end{tabular} \\
\hline

 RF &\begin{tabular}[c]{@{}c@{}}52.50\% \\ (+/- 5.30\%)\end{tabular} & \begin{tabular}[c]{@{}c@{}}56.50\% \\ (+/- 5.61\%)\end{tabular} & \begin{tabular}[c]{@{}c@{}}56.40\% \\ (+/- 3.61\%)\end{tabular} & \begin{tabular}[c]{@{}c@{}}54.60\% \\ (+/- 3.58\%)\end{tabular} & \begin{tabular}[c]{@{}c@{}}60.80\% \\ (+/- 4.42\%)\end{tabular} & \begin{tabular}[c]{@{}c@{}}66.50\% \\ (+/- 4.57\%)\end{tabular} & \begin{tabular}[c]{@{}c@{}}64.90\% \\ (+/- 4.41\%)\end{tabular} \\
\hline
\end{tabular} }
\label{tab:GRUValtset}
\end{table}

\begin{table}[]
\caption{{\color{black}P-values between GRU, ML techniques and SBS for the test set (750 instances) from RDS$_{\tau}$(1,2) in terms of Falkenauer's performance and number of bins achieved by GRU, ML techniques and SBS using paired Wilcoxon signed-rank test  corrected for multiple comparisons with the Bonferroni method}}
\label{tab:Pval-RDS(1,2)}
\centering
\begin{tabular}{c|ccccccc}
\hline

 & 0 & 0.01 & 0.02 & 0.03 & 0.04 & 0.05 & 0.06  \\
 \hline
 \multicolumn{7}{c}{\textbf{RDS$_{\tau}$1 Falkenauer’s Performance }} \\
\hline
GRU-SBS & 0.0044 & 0.0018 & 0.0027 & 0.0007 & 10$^{-07}$ & 10$^{-05}$ & 10$^{-08}$ \\
NN-SBS & 0.0136 & 0.0005 & 0.0002 & 0.0698 & 0.0001 & 10$^{-08}$ & 10$^{-09}$ \\
RF-SBS & 10$^{-05}$ & 0.0004 & 0.0036 & 0.0163 & 0.0006 & 0.0005 & 10$^{-07}$ \\
GRU-NN & 1.0000 & 1.0000 & 1.0000 & 0.8086 & 0.6112 & 0.0854 & 0.7340 \\
GRU-RF & 1.0000 & 1.0000 & 1.0000 & 1.0000 & 0.4460 & 1.0000 & 1.0000 \\
NN-RF & 0.3129 & 1.0000 & 0.9933 & 1.0000 & 1.0000 & 0.0019 & 0.0584 \\
 \hline
 \multicolumn{7}{c}{\textbf{RDS$_{\tau}$1 Bins }} \\
\hline
 GRU-SBS &  0.0022 & 0.0004 & 10$^{-06}$ & 10$^{-05}$ & 10$^{-10}$ & 10$^{-06}$ & 10$^{-08}$ \\
NN-SBS & 0.0077 & 0.0002 & 10$^{-08}$ & 0.0007 & 10$^{-06}$ & 10$^{-10}$ & 10$^{-12}$ \\
 RF-SBS & 10$^{-05}$ & 10$^{-06}$ & 10$^{-08}$ & 10$^{-05}$ & 10$^{-05}$ & 10$^{-05}$ & 10$^{-10}$ \\
GRU-NN & 1.0000 & 1.0000 & 1.0000 & 1.0000 & 0.3457 & 0.3458 & 0.0963 \\
GRU-RF & 1.0000 & 1.0000 & 1.0000 & 1.0000 & 0.4333 & 1.0000 & 1.0000 \\
NN-RF & 0.6745 & 0.7310 & 1.0000 & 1.0000 & 1.0000 & 0.0588 & 0.0501 \\
 
\hline
 \multicolumn{7}{c}{\textbf{RDS$_{\tau}$2 Falkenauer’s Performance }} \\
\hline
GRU-SBS & 10$^{-09}$ & 10$^{-08}$ & 10$^{-12}$ & 10$^{-08}$ & 0.0921 &10$^{-07}$ & 0.0001  \\
NN-SBS & 10$^{-10}$ & 10$^{-09}$ & 0.0006 & 10$^{-08}$ & 0.3837 & 1 & 1 \\
RF-SBS & 10$^{-10}$ & 10$^{-11}$ & 10$^{-08}$ & 10$^{-08}$ & 0.4030 & 0.0212 & 1 \\
GRU-NN & 1 & 1 & 10$^{-05}$ & 1 & 1 & 10$^{-08}$ & 0.0002 \\
GRU-RF & 1 & 0.7018 & 0.8429 & 1 & 1 & 0.0017 & 10$^{-05}$ \\
NN-RF & 1 & 1 & 10$^{-05}$ & 1 & 1 & 0.0031 & 1 \\
 \hline
 \multicolumn{7}{c}{\textbf{RDS$_{\tau}$2 Bins }} \\
\hline
GRU-SBS & 10$^{-07}$ & 10$^{-06}$ & 10$^{-08}$ & 10$^{-05}$ & 0.2596 & 10$^{-06}$ & 0.0024 \\
NN-SBS & 10$^{-09}$ & 10$^{-06}$ & 0.0094 & 10$^{-06}$ & 1 & 0.0304 & 1 \\
RF-SBS  & 10$^{-08}$ & 10$^{-08}$ & 10$^{-05}$ & 10$^{-05}$ & 0.8495 & 1 & 0.7293 \\
GRU-NN  & 0.8254 & 1 & 0.0006 & 1 & 1 & 10$^{-18}$ & 10$^{-05}$ \\
GRU-RF & 1 & 1 & 0.6180 & 1 & 1 & 10$^{-06}$ & 10$^{-08}$ \\
NN-RF & 0.7694 & 1 & 0.0164 & 1 & 1 & 10$^{-08}$ & 0.5059 \\
\hline
\end{tabular} 
\end{table}

\end{appendices}

\end{document}